%% file: main.tex
\documentclass{article}
\usepackage{iclr2018_conference,times}
\usepackage[utf8]{inputenc}
\usepackage[T1]{fontenc}
\usepackage{enumitem, amssymb, amsmath}
\usepackage{graphicx}
\usepackage{listings}
\usepackage{subfigure}
\usepackage{microtype}
\usepackage{float}
\usepackage{xcolor}
\usepackage{bm}

\usepackage{hyperref}       %
\usepackage{url} %

\hypersetup{
    colorlinks,
    linkcolor={red!50!black},
    citecolor={blue!50!black},
    urlcolor={blue!80!black}
}

\newlist{todolist}{itemize}{2}
\setlist[todolist]{label=$\square$}
\usepackage{pifont}

\newcommand{\addhspacesmall}{\hspace{1pt}}

\newcommand{\norm}[1]{\left\lVert#1\right\rVert}

\usepackage[]{authblk}

\newcommand{\todo}[1]{\textcolor{red}{#1}}

 \renewcommand{\todo}[1]{}  %

\newcommand\blfootnote[1]{%
  \begingroup
  \renewcommand\thefootnote{}\footnote{#1}%
  \addtocounter{footnote}{-1}%
  \endgroup
}

\input{math_commands}

\iclrfinalcopy

\title{Many Paths to Equilibrium: GANs Do Not Need to Decrease a Divergence At Every Step}

\author[1]{William Fedus$^*$ }
\author[2]{Mihaela Rosca$^*$ }
\author[2]{Balaji Lakshminarayanan}%
\author[1]{\authorcr Andrew M. Dai}%
\author[2]{Shakir Mohamed}%
\author[1]{Ian Goodfellow}%
\affil[1]{Google Brain}
\affil[2]{DeepMind}

\begin{document}

\maketitle

\begin{abstract}
Generative adversarial networks (GANs)
are a family of generative models that do not minimize a single
training criterion.
Unlike other generative models, the data distribution is learned via a game
between a generator (the generative model) and a discriminator (a teacher
providing training signal) that each minimize their own cost.
GANs are designed to reach a Nash equilibrium at which each player cannot
reduce their cost without changing the other players' parameters.
One useful approach for the theory of GANs is to show that a divergence
between the training distribution and the model distribution obtains its
minimum value at equilibrium.
Several recent research directions have been motivated by the idea that
this divergence is the primary guide for the learning process and that
every step of learning should decrease the divergence.
We show that this view is overly restrictive.
During
GAN training, the discriminator provides learning signal in situations where
the gradients of the divergences between distributions would
not be useful.
We provide empirical counterexamples to the view of GAN training as
divergence minimization.
Specifically, we demonstrate
that GANs are able to learn
distributions in situations where the divergence minimization point of view
predicts they would fail.
We also show that gradient
penalties motivated from the divergence minimization perspective
are equally helpful when applied in other contexts in which the
divergence minimization perspective does not predict they would be helpful.
This contributes to a growing body of evidence that GAN training may be
more usefully viewed as approaching Nash equilibria via trajectories that
do not necessarily minimize a specific divergence at each step.
\end{abstract}

\blfootnote{$^*$ Equal contribution.}
\vspace{-1cm}
\section{Introduction}

Generative adversarial networks (GANs) \citep{gan}
are generative models based on a competition between a generator
network $G$ and a discriminator network $D$.
The generator network $G$ represents a probability distribution
$\pmodel(\vx)$. To obtain a sample from this distribution, we apply
the generator network to a noise vector $\vz$ sampled from $p_\vz$, that is $\vx = G(\vz)$.
Typically, $\vz$ is drawn from a Gaussian or uniform distribution,
but any distribution with sufficient diversity is possible.
The discriminator $D(\vx)$ attempts to distinguish whether an input
value $\vx$ is real (came from the training data) or fake (came from
the generator).

The goal of the training process is to recover the true distribution
$\pdata$ that generated the data.
Several variants of the GAN training process have been proposed.
Different variants of GANs have been interpreted as approximately
minimizing different divergences or distances between $\pdata$ and
$\pmodel$.
However, it has been difficult to understand
whether the improvements are caused by a change in the underlying divergence or the learning dynamics.

We conduct several experiments to assess whether  the improvements associated
with new GAN methods are due to the reasons cited in their design motivation.
We perform a comprehensive study of GANs on simplified, synthetic tasks for which
the true $\pdata$ is known and the relevant distances are straightforward to
calculate, to assess the performance of proposed models against baseline methods.
 We also evaluate GANs using several independent evaluation measures on real data
to better understand new approaches. Our contributions are:
\begin{itemize}

\item  We aim to clarify terminology used in recent papers, where the terms
``standard GAN,'' ``regular GAN,'' or ``traditional GAN'' are used without
definition (e.g., \citep{wgan, denton2015deep, improvedgan, donahue2016adversarial}).
The original GAN paper described two different losses: the ``minimax'' loss and
the ``non-saturating'' loss, equations (10) and (13) of
\citet{goodfellow2016tutorial}, respectively.
Recently, it has become important to clarify this terminology,
because many of the criticisms of ``standard GANs'', e.g. \citet{wgan},
are applicable only to the minimax GAN,
while the non-saturating GAN is the standard for GAN implementations.
The non-saturating GAN was
recommended for use in practice and implemented in
the original paper of \citet{gan},
and is the default in subsequent papers
\citep{dcgan, improvedgan, donahue2016adversarial, nowozin2016f}\footnote{
	The original GAN paper implements both the minimax and non-saturating cost
	but uses the non-saturating cost for the published configurations of
	experiments:
	\url{https://github.com/goodfeli/adversarial/blob/master/cifar10_convolutional.yaml\#L139}.
	To the best of our knowledge, the DCGAN codebase implements {\em only} the
	non-saturating cost:
	\url{https://github.com/soumith/dcgan.torch/blob/master/main.lua\#L215}.
	Likewise, the improved-gan codebase implements {\em only} the
	non-saturating cost:
	\url{https://github.com/openai/improved-gan/blob/master/imagenet/build_model.py\#L114}.
	If only one of these two costs were to be called ``standard,'' it should be the
	non-saturating version.
}.
To avoid confusion we will always indicate whether we mean minimax GAN (M-GAN) or
non-saturating GAN (NS-GAN).

\item We demonstrate that gradient penalties designed in the divergence minimization framework---to
improve Wasserstein GANs \citep{wgangp} or justified from a game theory perspective
to improve minimax GANs \citep{dragan}---also improve the non-saturating GAN on both synthetic and real data. We observe improved sample quality and diversity.

\item We find that non-saturating GANs are able to fit problems
  that cannot be fit by Jensen-Shannon divergence
	minimization.
  Specifically, Figure~\ref{fig:parallel_lines} shows a GAN
  using the loss from the original non-saturating GAN succeeding on a task where
  the Jensen-Shannon divergence provides no useful gradient. Figure~\ref{fig:no_vanish} shows that the non-saturating GAN does not suffer from
  vanishing gradients when applied to two widely separated Gaussian distributions.

\end{itemize}

\section{Variants of Generative Adversarial Networks}

\subsection{Non-Saturating and Minimax GANs}
In the original GAN formulation \citep{gan},
the output of the discriminator is a probability and
the cost  function for the discriminator is given by
the negative log-likelihood of the binary discrimination task of classifying
samples as real or fake:

\begin{equation}
J^{(D)}(D,G) = - \underset{ \vx \sim \pdata } {\E}  \left[\log D(\vx)\right]  - \underset{ \vz \sim p_\vz } {\E}  \left[\log (1 - D(G(\vz))) \right].
\label{eq:disc}
\end{equation}

The theoretical analysis in \citep{gan} is based on a zero-sum
game in which the generator maximizes $J^{(D)}$, a situation that we refer to here
as ``minimax GANs''.
In minimax GANs the generator attempts to generate samples that have low probability of being fake,
by minimizing the objective (\ref{eq:gen_loss_minimax}).
However, in practice, \citet{gan} recommend implementing
an alternative cost function that instead ensures that generated samples have
high probability of being real, and the generator instead minimizes
an alternative objective (\ref{eq:gen_loss}).
\begin{eqnarray}
\textrm{\textbf{Minimax}} & J^{(G)}(G) =&  \underset{ \vz \sim p_\vz } {\E} \log [1 - D(G(\vz))].
\label{eq:gen_loss_minimax}\\
\textrm{\textbf{Non-saturating}} & J^{(G)}(G) =& - \underset{ \vz \sim p_\vz } {\E} \log D(G(\vz)).
\label{eq:gen_loss}
\end{eqnarray}
We refer to the alternative objective as non-saturating, due to the
non-saturating behavior of the gradient (see figure \ref{fig:no_vanish}), and
was the implementation used in the code of the original paper.
We use the non-saturating objective (\ref{eq:gen_loss}) in all our experiments

As shown in \citep{gan}, whenever $D$ successfully minimizes
$J^{(D)}$ optimally, maximizing $J^{(D)}$ with respect to the generator
is equivalent to minimizing the Jensen-Shannon divergence.
\citet{gan} use this observation to establish that there is
a unique Nash equilibrium in function space corresponding to
$\pdata = \pmodel$.

\subsection{Wasserstein GAN}
Wasserstein GANs \citep{wgan} modify the discriminator to emit an unconstrained
real number rather than a probability
(analogous to emitting the logits rather than the probabilities used in the original GAN paper).
The cost function for the WGAN then omits the log-sigmoid functions used in the original
GAN paper. The cost function for the discriminator is now:
\begin{equation}
    W^{(D)}(D,G) =  \underset{ \vx \sim \pdata } {\E}   \left[D(\vx)\right]
    - \underset{ \vz \sim p_\vz } {\E}  \left[D(G(\vz)) \right].
\end{equation}
The cost function for the generator is simply $W^{(G)}=-W^{(D)}(D,G)$.
When the discriminator is Lipschitz smooth, this approach approximately minimizes
the earth mover's distance between $\pdata$ and $\pmodel$.
To enforce Lipschitz smoothness, the weights of $D$ are clipped to lie within $(-c, c)$ where $c$ is some small real number.

\subsection{Gradient Penalties for Generative Adversarial Networks}
\label{sec:grad_penalties}
Multiple formulations of gradient penalties have been proposed for GANs. As
introduced in WGAN-GP \citep{wgangp}, the gradient penalty is justified from the
perspective of the Wasserstein distance, by imposing properties which hold for
an optimal critic as an additional training criterion. In this approach, the
gradient penalty is typically a penalty on the gradient norm, and is applied on
a linear interpolation between data points and samples, thus smoothing out the
space between the two distributions.

\citet{dragan} introduce DRAGAN with a gradient penalty from the perspective of regret minimization, by setting the regularization function to be a gradient penalty on points around the data manifold, as in \emph{Follow The Regularized Leader} \citep{cesa2006prediction}, a standard no-regret algorithm.
This encourages the discriminator to be close to linear around the data manifold, thus bringing the set of possible discriminators closer to a convex set, the set of linear functions.  We also note that they used the minimax version of the game to define the loss, in which the generator maximizes $J^{(D)}$ rather than minimizing $J^{(G)}$.

To formalize the above, both proposed gradient penalties of the form:

\begin{equation}
  \underset{ \hat{x} \sim \phatx } {\E}  \left[ ( \norm{\nabla_{ \hat{x} } D(\hat{x})}_2 - 1 )^2 \right],
\end{equation}

where $\phatx$ is defined as the distribution defined by the sampling process:
\begin{eqnarray}
&	x  \sim \pdata; \qquad
 	x_\textrm{model}  \sim \pmodel; \qquad
	x_\textrm{noise}  \sim \pnoise
		\end{eqnarray}
\begin{subequations}
	\begin{align}
	\textrm{\textbf{DRAGAN}} \quad & \tilde{x} = x + x_\textrm{noise} \\
   \textrm{\textbf{WGAN-GP}}\quad & \tilde{x} = x_\textrm{model}
	\end{align}
\end{subequations}
\begin{align}
	&	\alpha   \sim U(0, 1) \\
	\hat{x} &= \alpha x + (1 - \alpha) \tilde{x}.
\end{align}

As we will note in our experimental section, \citet{dragan} also reported
that mode-collapse is reduced using their version of the gradient penalty.

\subsubsection{Non-saturating GAN with Gradient Penalty}

We consider the non-saturating GAN objective (\ref{eq:gen_loss}) supplemented by
two gradient penalties: the penalty proposed by \citet{wgangp}, which we refer
to as ``GAN-GP''; the gradient penalty proposed by DRAGAN \citep{dragan}, which
we refer to as DRAGAN-NS, to emphasize that we use the non-saturating generator
loss function. In both cases, the gradient penalty applies only to the
discriminator, with the generator loss remaining unchanged (as defined in
Equation~\ref{eq:gen_loss}). In this setting, the loss of the discriminator
becomes:
\begin{equation}
  \tilde{J}^{(D)}(D,G) =  -\underset{ x \sim \pdata } {\E }  \left[\log D(x)\right]  - \underset{ z \sim p_z } {\E}  \left[\log (1 - D(G(z))) \right]  + \lambda  \underset{ \hat{x} \sim \phatx} {\E}  \left[ ( \| \nabla_{ \hat{x} } D(\hat{x}) \|_2 - 1 )^2 \right]
\label{eq:discgp}
\end{equation}

We consider these GAN variants because:
\begin{itemize}
\item We want to assess whether gradient penalties are effective outside their original defining scope. Namely, we perform experiments to determine whether the benefit obtained by applying the gradient penalty for Wasserstein GANs is obtained from properties of the earth mover's distance, or from the penalty itself. Similarly, we evaluate whether the DRAGAN gradient penalty is beneficial outside the minimax GAN setting.
\item We want to assess whether the exact form of the gradient penalty matters.
\item We compare three models, to control over different aspects of training: same gradient penalty but different underlying adversarial losses (GAN-GP versus WGAN-GP), as well as the same underlying adversarial loss, but different gradient penalties (GAN-GP versus DRAGAN-NS).
\end{itemize}

We note that we do not compare with the original DRAGAN formulation, which uses the minimax GAN formulation, since in this work we focus on non-saturating GAN variants.

\begin{figure}%
\begin{center}
    \begin{subfigure}[Step 0]{\includegraphics[width=3.5cm]{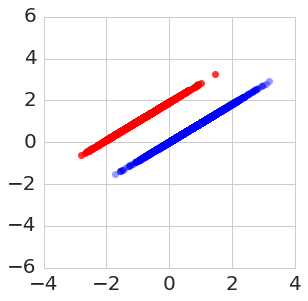} }%
    \end{subfigure}
    \begin{subfigure}[Step 5000]{\includegraphics[width=3.5cm]{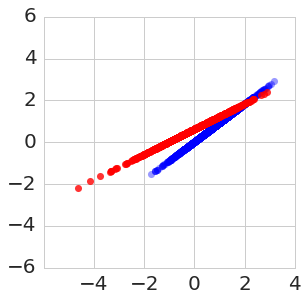} }%
    \end{subfigure}
    \begin{subfigure}[Step 12500]{\includegraphics[width=3.5cm]{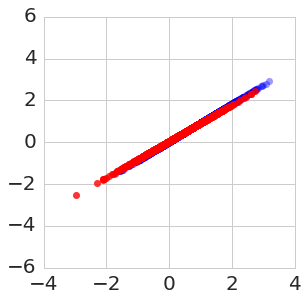} }%
    \end{subfigure}
    \caption{Visualization of experiment 1 training dynamics in two dimensions,
      demonstrated specifically in the case where the model is initialized so
      that it it represents a linear manifold parallel to the linear manifold
      of the training data.
    Here the GAN model (red points) converges upon the one dimensional
  synthetic data distribution (blue points).
Specifically, this is an illustration of the parallel line thought experiment
from \citep{wgan}.
When run in practice with a non-saturating GAN, the GAN succeeds.
In the same setting, minimization of Jensen-Shannon divergence would fail.
This indicates that while Jensen-Shannon divergence is useful for
characterizing GAN equilibrium, it does not necessarily tell us much about
non-equilibrium learning dynamics.}
    \label{fig:parallel_lines}
\end{center}
\end{figure}

\begin{figure}[h]
  \begin{center}
    \includegraphics[width=0.7\textwidth]{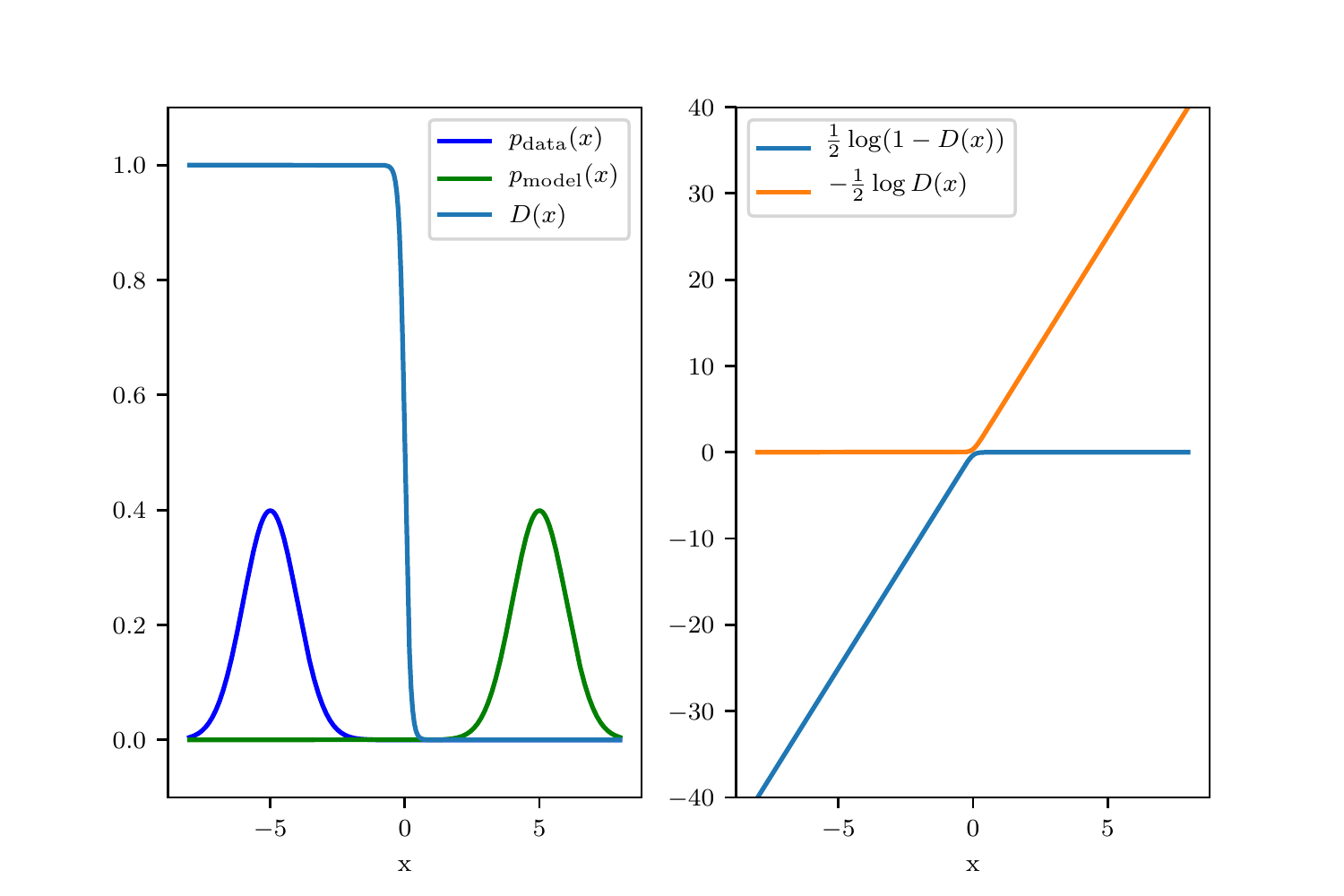}
    \caption{
      (Left) A recreation of Figure 2 of \citet{wgan}.
      This figure is used
      by \citet{wgan} to show that a model they call the ``traditional GAN'' suffers
      from vanishing gradients in the areas where $D(x)$ is flat. This plot is correct
      if ``traditional GAN'' is used to refer to the {\em minimax} GAN, but it does not
      apply to the {\em non-saturating} GAN.
      (Right) A plot of both generator losses from the original GAN paper, as a
      function of the generator output.
      Even when the model distribution is highly separated from
      the data distribution, non-saturating GANs
      are able to bring the model
      distribution closer to the data distribution
      because the {\em loss function} has strong gradient
      when the generator samples are far from the data samples,
      {\em even when the discriminator itself has nearly zero gradient}.
      While it is true that the
      $\frac{1}{2} \log( 1 - D(x))$ loss has a vanishing gradient on the right half
      of the plot, the original GAN paper instead recommends implementing
      $-\frac{1}{2} \log D(x)$. This latter, recommended loss function has a vanishing
      gradient only on the left side of the plot. It makes sense for the gradient to
      vanish on the left because generator samples in that area have already reached
the area where data samples lie.
  }
  \label{fig:no_vanish}
\end{center}
\end{figure}

\section{Many Paths to Equilibrium}

The original GAN paper \citep{gan} used the correspondence between $J^{(D)}(D^*, G)$ and the
Jensen-Shannon divergence to characterize the {\em Nash equilibrium} of minimax GANs.
It is important to keep in mind that there are many ways for the learning
process to approach this equilibrium point, and the majority of them do not
correspond to gradually reducing the Jensen-Shannon divergence at each step.
Divergence minimization is useful for understanding the outcome
of training, but GAN training is not the same thing as running gradient descent
on a divergence and GAN training may not encounter the same problems as gradient
descent applied to a divergence.

\citet{wgan} describe the {\em learning process} of GANs from the perspective
of divergence minimization and show that the Jensen-Shannon divergence is
unable to provide a gradient that will bring $\pdata$ and $\pmodel$ together if
both are sharp manifolds that do not overlap early in the learning process.
Following this line of reasoning, they suggest that when applied to
probability distributions that are supported only on low dimensional
manifolds, the Kullback Leibler (KL), Jensen Shannon (JS) and Total
Variation (TV) divergences do not provide a useful gradient for learning
algorithms based on gradient descent, the ``traditional GANs'' is inappropriate for
fitting such low dimensional manifolds (``traditional GAN'' seems to refer to
the minimax version of GANs used for theoretical analysis in the original paper,
and there is no explicit statement about whether the argument is intended to
apply to the non-saturating GAN implemented in the code accompanying the original
GAN paper). In Section~\ref{sec:synth_exps} we show that non-saturating GANs
are able to learn on tasks where the data distribution lies on the low dimensional
manifold.

We show that non-saturating GANs do not suffer from vanishing gradients for
two widely separated Gaussians in Figure~\ref{fig:no_vanish}.
The fact that the gradient of the recommended loss does not actually
vanish explains why GANs with the non-saturating objective (\ref{eq:gen_loss}),
are able to bring together two widely separated Gaussian distributions.
Note that the gradient for this loss does not vanish {\em even when the discriminator
is optimal}.
The {\em discriminator} has vanishing gradients but the {\em generator loss}
amplifies small differences in discriminator outputs to recover strong gradients.
This means it is possible to train the GAN by changing the {\em loss} rather than
the discriminator.

For the parallel lines thought experiment \citep{wgan} (see Figure~\ref{fig:parallel_lines}),
the main problem with the Jensen-Shannon divergence is that it is parameterized
in terms of the density function, and the two density functions have no support in common.
Most GANs, and many other models, can solve this problem by parameterizing their loss functions
in terms of samples from the two distributions rather than in terms of their density functions.
\section{Synthetic Experiments}
\label{sec:synth_exps}

To assess the learning process of GANs we empirically examine GAN training on pathological tasks where the data is constructed to lie on a low dimensional manifold, and show the model is able to learn the data distribution in cases where using the underlying divergence obtained at optimality would not provide useful gradients. We then evaluate convergence properties of common GAN variants on this task where the parameters generating the distribution are known.

\subsection{Experiment I:  1-D Data Manifold and 1-D generator}
In our first experiment, we generate synthetic training data that lies along a one-dimensional line and design a one-dimensional generative model, however, we embed the problem in a higher $d$-dimensional space where $d\gg 1$.
This experiment is essentially an implementation of a thought experiment from
\citet{wgan}.

Specifically, in a $d$-dimensional space, we define $\pdata$ by randomly generating
parameters defining the distribution once at the beginning of the experiment.
We generate a random $b_r \in \R^d$ and random $W_r \in \R^{1 \times d}$.
Our latent $z_r \in \R \sim N(0,\sigma)$ where $\sigma$ is the standard deviation of the normal distribution.  The synthetic training data of $m$ examples is then given by

\begin{equation} \label{eq:one_d_data}
    \{x^{(i)}\}_{i=1}^{m} = \{z_r^{(i)}\}_{i=1}^{m} W_r + b_r
\end{equation}

\noindent The real synthetic data is therefore Gaussian distributed on a 1-D surface within the space, where the position is determined by $b_r$ and the orientation is determined by $W_r$.

The generator also assumes the same functional form, that is, it is also intrinsically one dimensional,

\begin{equation}
G_{\theta}(z) = z W_{\theta} + b_{\theta}
\end{equation}

\noindent where $b_{\theta} \in \R^d$ and
$W_{\theta} \in \R^{1 \times d}$.
The discriminator is a single hidden layer ReLU network, which is of higher
complexity than the generator so that it may learn non-linear boundaries in
the space.

This experiment captures the idea of sharp, non-overlapping manifolds that motivate alternative GAN losses.  Further, because we know the true generating parameters of the training data, we may explicitly test convergence properties of the various methodologies.

\subsection{Experiment II: 1-D Data Manifold and overcomplete generator}
In our second experiment, the synthetic training data is still the same (lying on a 1-D line) and given by Eq. \ref{eq:one_d_data} but now the generator is overcomplete for this task, and has a higher latent dimension $g$, where $1 < g \leq d$.

\begin{equation}
G(z) = z W_{\theta} + b_{\theta}
\end{equation}

\noindent where matrix $W_{\theta} \in \R^{g \times d}$ and vector
$b_{\theta} \in \R^d$, so that the generator is able to represent a manifold
with too high of a dimensionality.
The generator parameterizes a multivariate Gaussian $N(x; \mu, \Sigma$) with $\mu = b$.  The covariance matrix elements $\Sigma_{ij} = E[\sigma^2 (X_i - \mu_i)(X_j - \mu_j)] = \sigma^2 E[(X_i - \mu_i)(X_j - \mu_j)]$.  In vector notation,  $\Sigma = \sigma^2 W^T W$.

\subsection{Results}
To evaluate the convergence of an experimental trial, we report the square {F}r\'{e}chet distance (\citet{frechet}) between the true data Gaussian distribution and the fitted Gaussian parameters.
In our notation, where the $r$ subscript
denotes real data, the $\theta$ subscript denotes the generator Gaussian
parameters and $\lVert x \rVert ^ 2$ is the squared $l_2$ norm of $x$, the {F}r\'{e}chet distance is defined as (\citet{frechet_multivariate_gasussians}):

\begin{equation}
d^2(\mu_{r},\mu_{\theta},\Sigma_{r},\Sigma_{\theta}) = \lVert \mu_{\theta} - \mu_r \rVert ^ 2 + \Tr\left(\Sigma_r + \Sigma_{\theta} - 2 (\Sigma_r \Sigma_{\theta})^{-1/2}\right)
\end{equation}

Every GAN variant was trained for 200000 steps. For each step, the generator is updated once and the discriminator is updated 5 times.  Throughout the paper, the number of steps will correspond to the number of generator updates.

\begin{figure}[h]%
\begin{center}
  \begin{subfigure}[Non-saturating GAN training at 0, 10000 and 20000 steps.]{
    \includegraphics[width=3cm]{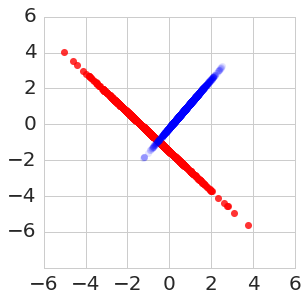} \addhspacesmall %
    \includegraphics[width=3cm]{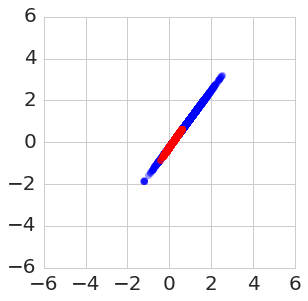} \addhspacesmall %
    \includegraphics[width=3cm]{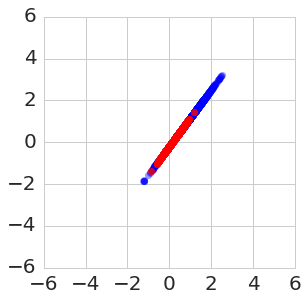} \addhspacesmall%
    }
    \end{subfigure}

  \vspace{3mm}
    \begin{subfigure}[GAN-GP training at 0, 10000 and 20000 steps.]{
    \includegraphics[width=3cm]{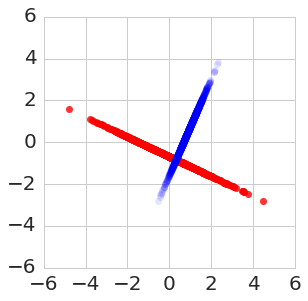} \addhspacesmall %
    \includegraphics[width=3cm]{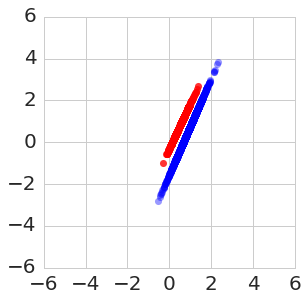} \addhspacesmall %
    \includegraphics[width=3cm]{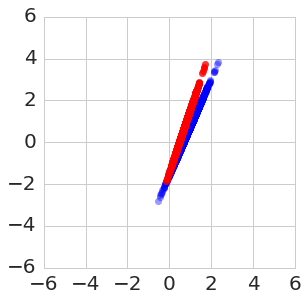} \addhspacesmall%
    }
    \end{subfigure}

  \vspace{3mm}
    \begin{subfigure}[DRAGAN-NS training at 0, 10000 and 20000 steps.]{
    \includegraphics[width=3cm]{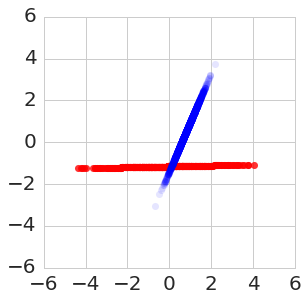} \addhspacesmall %
    \includegraphics[width=3cm]{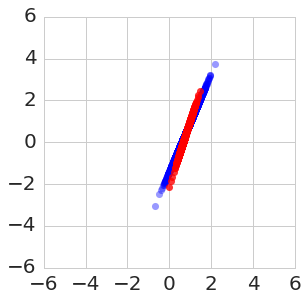} \addhspacesmall %
    \includegraphics[width=3cm]{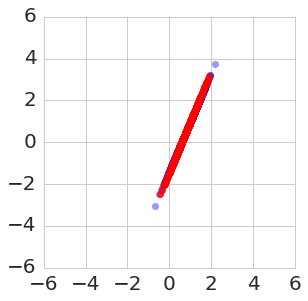} \addhspacesmall%
    }
    \end{subfigure}

    \caption{Visualization of experiment 1 training dynamics in two dimensions.
    Here the GAN model (red points) converges upon the one dimensional
  synthetic data distribution (blue points). We note that this is a visual
illustration, and the results have not been averaged out over multiple seeds.
Exact plots may vary on different runs.
However, a single example of success is sufficient to refute claims that this
this task is impossible for this model.
}
    \label{fig:training_dynamics_1}
\end{center}
\end{figure}

\begin{figure}[h]%
\begin{center}
  \begin{subfigure}[Non-saturating GAN training at 0, 5000 and 10000 steps.]{
    \includegraphics[width=3cm]{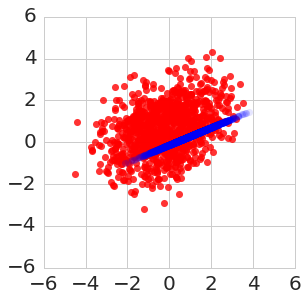} \addhspacesmall %
    \includegraphics[width=3cm]{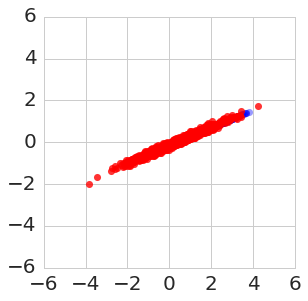} \addhspacesmall %
    \includegraphics[width=3cm]{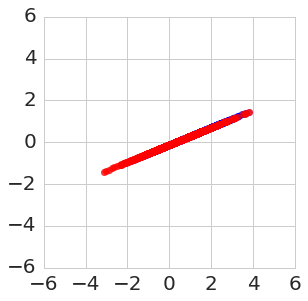} \addhspacesmall %
    }
    \end{subfigure}

  \vspace{3mm}
    \begin{subfigure}[GAN-GP training at 0, 5000 and 10000 steps.]{
    \includegraphics[width=3cm]{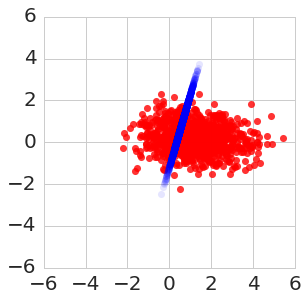} \addhspacesmall %
    \includegraphics[width=3cm]{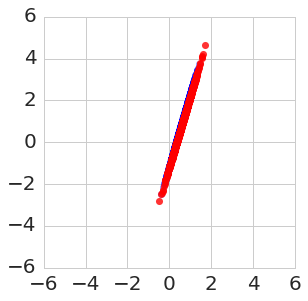} \addhspacesmall %
    \includegraphics[width=3cm]{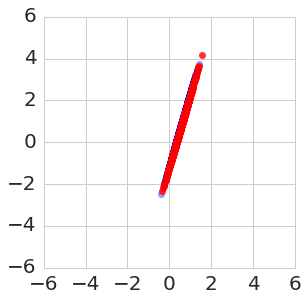} \addhspacesmall %
    }
    \end{subfigure}

  \vspace{3mm}
    \begin{subfigure}[DRAGAN-NS training at 0, 5000 and 10000 steps]{
    \includegraphics[width=3cm]{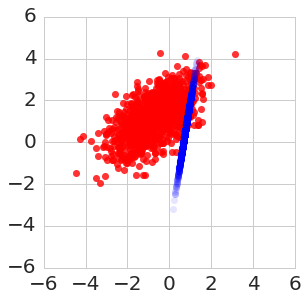} \addhspacesmall %
    \includegraphics[width=3cm]{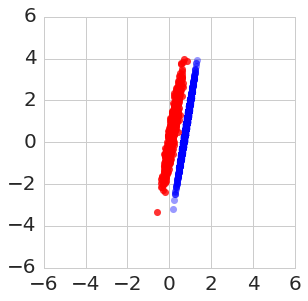} \addhspacesmall %
    \includegraphics[width=3cm]{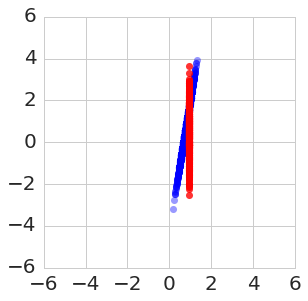} \addhspacesmall %
    }
    \end{subfigure}

    \caption{Visualization of experiment 2 training dynamics in two dimensions - where the GAN model has 3 latent variables. Here the rank one GAN model (red points) converges upon the one dimensional synthetic data distribution (blue points).
    We observe how for poor initialization the non-saturating GAN
suffers from mode collapse.
However, adding a gradient penalty stabilizes training. We note that this is a visual illustration, and the results have not been averaged out over multiple seeds. Exact plots may vary on different runs.}
    \label{fig:training_dynamics_2}
\end{center}
\end{figure}

The main conclusions from our synthetic data experiments are:
\begin{itemize}
  \item Gradient penalties (both applied near the data manifold, DRAGAN-NS, and at
  an interpolation between data and samples, GAN-GP) stabilize training and
  improve convergence (Figures~\ref{fig:training_dynamics_1},
  ~\ref{fig:hyper_exp1_lr}, ~\ref{fig:hyper_exp2_input_dims}).

  \item Despite the inability of Jensen-Shannon divergence minimization
    to solve this problem, we find that the non-saturating GAN succeeds in converging to the
  1D data manifold (Figure~\ref{fig:training_dynamics_1}).  However, in higher
  dimensions the resulting fit is not as strong as the other methods: Figure~\ref{fig:hyper_exp2_input_dims} shows that increasing the number of dimensions while keeping the learning rate fixed can decrease the performance of the non-saturating GAN model.
  \item Non-saturating GANs are able to learn data distributions which are disjoint
    from the training sample distribution at initialization (or another point
    in training), as demonstrated in Figure~\ref{fig:parallel_lines}.
  \item Updating the discriminator 5 times per generator update does not result in
    vanishing gradients when using the non-saturating cost. However, when scaling the number of discriminator updates to 100 per generator update, non-saturating GANs perform worse than when using a smaller number of updates (1, 5, 10). Gradient penalties help here too: GAN-GP scales better with the number of discriminator updates. The results are detailed in Appendix Section~\ref{sec:synth_updates}.
  \item An over-capacity generator with the ability to have more directions of high
    variance than the
  underlying data is able to capture the data distribution using non-saturating GAN
  training (Figure~\ref{fig:training_dynamics_2}).

\end{itemize}

\subsection{Hyperparameter sensitivity}

We assess the robustness of the considered models by looking at results across hyperparameters for both experiment 1 and experiment 2. In one setting, we keep the input dimension fixed while varying the learning rate (Figure~\ref{fig:hyper_exp1_lr}); in another setting, we keep the learning rate fixed, while varying the input dimension (Figure~\ref{fig:hyper_exp2_input_dims}). In both cases, the results are averaged out over 1000 runs per setting, each starting from a different random seed. We notice that:
\begin{itemize}
  \item The non-saturating GAN model (with no gradient penalty) is most sensitive to hyperparameters.
  \item Gradient penalties make the non-saturating GAN model more robust.
\item Both Wasserstein GAN formulations are quite robust to hyperparameter changes.
\item For certain hyperparameter settings, there is no performance difference between
  the two gradient penalties for the non-saturating GAN, when averaging across random seeds.
  This is especially visible in Experiment 1, when the number of latent variables is 1.
  This could be due to the fact that the data sits on a low dimensional
  manifold, and because the discriminator is a small, shallow network.
\end{itemize}

\begin{figure*}[!]
\begin{center}
\begin{subfigure}[Synthetic Experiment 1]{
\includegraphics[width=6.7cm]{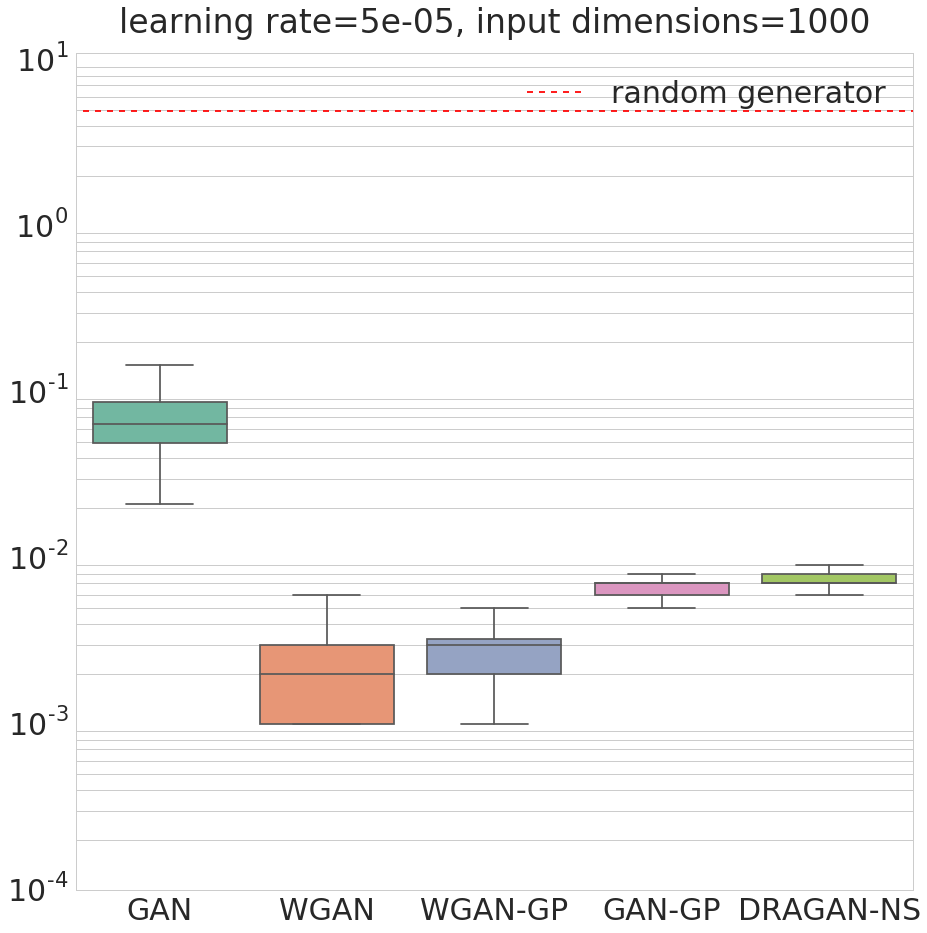} \addhspacesmall
\label{fig:fd_exp1}
}\end{subfigure}
\begin{subfigure}[Synthetic Experiment 2]{
\includegraphics[width=6.7cm]{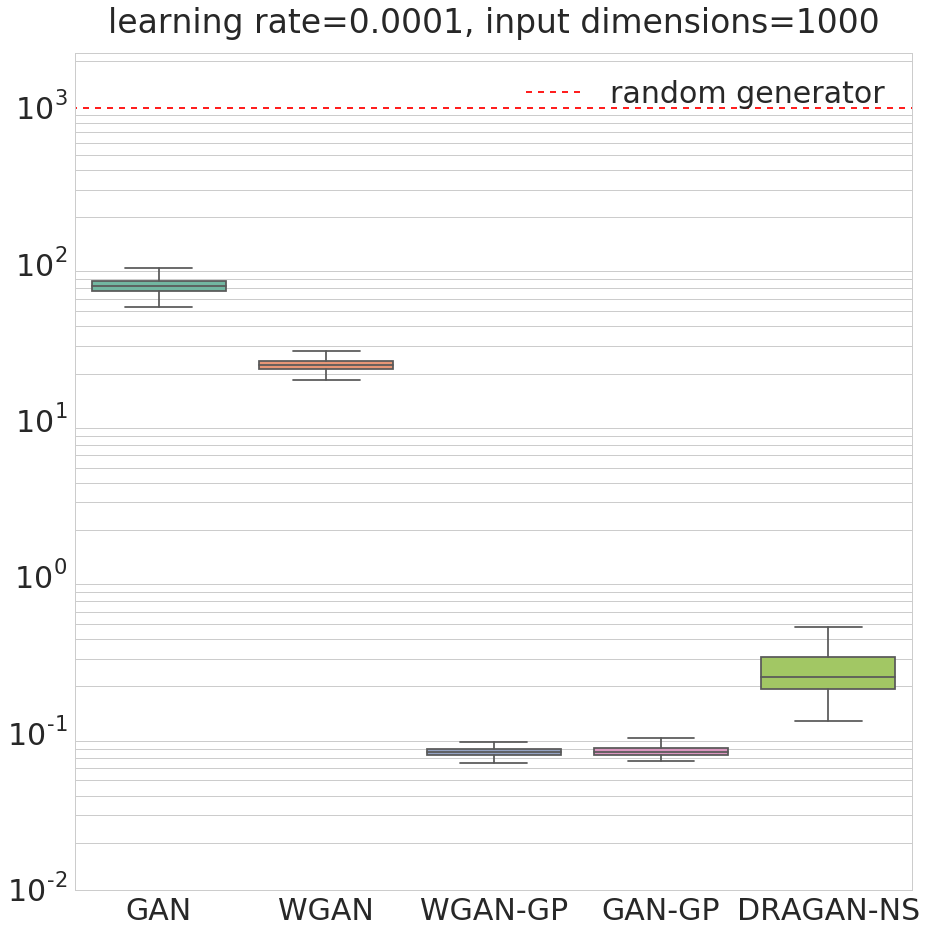}
\label{fig:fd_exp2}
}\end{subfigure}
\caption{{\textbf{The square {F}r\'{e}chet distance} between the learned Gaussian and the true Gaussian distribution. For reference, we also plot the distance obtained by a randomly initialized generator with the same architecture as the trained generators. Results are averaged over 1000 runs. Lower values are better.}
}
\label{fig:fd_synth}
\end{center}
\vspace{-4mm}
\end{figure*}

\section{Real data experiments}
\begin{figure*}
\begin{center}
\includegraphics[width=0.24\textwidth]{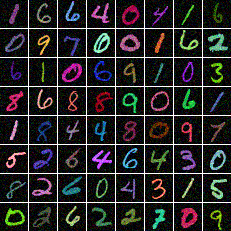} \addhspacesmall
\includegraphics[width=0.24\textwidth]{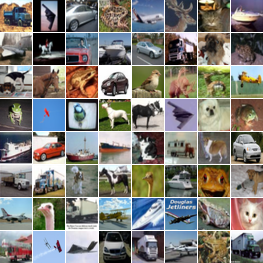} \addhspacesmall
\includegraphics[width=0.24\textwidth]{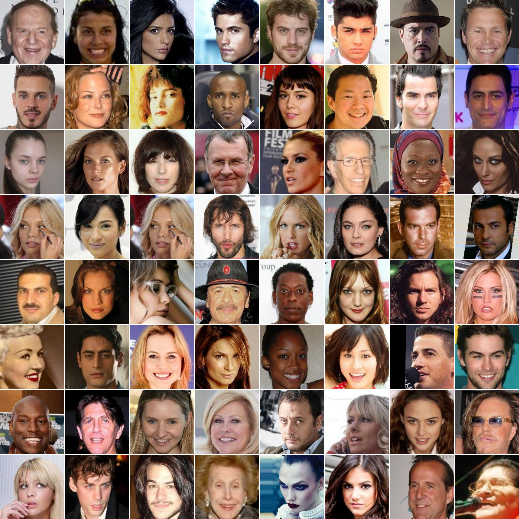} \addhspacesmall
\caption{Examples from the three datasets explored in this paper: Color MNIST (left), CIFAR-10 (middle) and CelebA (right).}
\label{fig:data_examples}
\end{center}
\end{figure*}

To assess the effectiveness of the gradient penalty on standard datasets for
the non-saturating GAN formulation, we train a
non-saturating GAN,
a non-saturating GAN with the gradient penalty introduced by \citep{wgangp} (denoted by GAN-GP),
a non-saturating GAN with the gradient penalty introduced by \citep{dragan} (denoted by DRAGAN-NS),
and a Wasserstein GAN with gradient penalty (WGAN-GP) on three datasets: Color MNIST \citep{unrolledgan} - data dimensionality $(28, 28, 3)$, CelebA \citep{celeba} - data dimensionality $(64, 64, 3)$ and CIFAR-10 \citep{cifar10} - data dimensionality $(32, 32, 3)$, as seen in Figure~\ref{fig:data_examples}.

For all our experiments we used $\lambda=10$ as the gradient penalty coefficient and used
batch normalization \citep{batchnorm}; \citet{dragan} suggests that batch normalization
is not neeeded for DRAGAN, but we found that it also improved our DRAGAN-NS results.
We used the Adam optimizer \citep{adam} with $\beta_1 = 0.5$ and $\beta_2 = 0.9$ and a
batch size of 64. The input data was scaled to be between -1 and 1.
We did not add any noise to the discriminator inputs or activations, as that
regularization technique can be interpreted as having the same goal as gradient penalties,
and we wanted to avoid
a confounding factor.  We trained all Color MNIST  models for 100000 steps, and CelebA and CIFAR-10 models for 200000 steps. We note that the experimental results on real data for the
non-saturating GAN and for the Improved Wasserstein GAN (WGAN-GP) are
quoted with permission from an earlier publication by
\citet{alphagan}.

We note that the WGAN-GP model was the only model for which we did 5 discriminator updates in real data experiments. All other models (DCGAN, DRAGAN-NS, GAN-GP) used one discriminator update for generator update.

For all reported results, we sweep over two hyperparameters:
\begin{itemize}
    \item Learning rates for the discriminator and generator. Following \citet{dcgan}, we tried learning rates of 0.0001, 0.0002, 0.0003 for both the discriminator and the generator. We note that this is consistent with WGAN-GP, where the authors use 0.0002 for CIFAR-10 experiments.
    \item Number of latents. For CelebA and CIFAR-10 we try latent sizes 100, 128 and 150, while for Color MNIST we try 10, 50, 75.
\end{itemize}

\subsection{Evaluation}

Unlike the synthetic case, here we are unable to evaluate the performance of our models relative to the true solution, since that is unknown. Moreover, there is no single metric that can evaluate the performance of GANs. We thus complement visual inspection with three metrics, each measuring a different criteria related to model performance. We use the Inception Score \citep{improvedgan} to measure how visually appealing CIFAR-10 samples are, the MS-SSIM metric \citep{wang2003multiscale, acgan} to check sample diversity, and an Improved Wasserstein independent critic to assess overfitting, as well as sample quality \citep{nvpgan}. For a more detailed explanation of these metrics, we refer to  \citet{alphagan}. In all our experiments, we control over discriminator and generator architectures, using the ones used by DCGAN \citep{dcgan} and the original WGAN paper \citep{wgan}\footnote{Code at: \url{https://github.com/martinarjovsky/WassersteinGAN/blob/master/models/dcgan.py}}. We note that the WGAN-GP paper used a different architecture when reporting the Inception Score on CIFAR10, and thus their results are not directly comparable.

For all the metrics, we report both the hyperparameter sensitivity of the model (by showing quartile statistics), as well as the 10 best results according to the metric. The sample diversity measure needs to be seen in context with the value reported on the test set: too high diversity can mean failure to capture the data distribution. For all other metrics,
higher is better.

\subsection{Visual sample inspection}

By visually inspecting the results of our models, we noticed that applying gradient penalties to the
non-saturating GAN results in more stable training across the board.
When training the non-saturating GAN with no gradient penalty, we did observe
cases of severe mode collapse (see Figure~\ref{fig:mode_collapse_standard_gan}).
Gradient penalties improves upon that, but we can still observe mode collapse. Each non-saturating GAN variant with gradient penalty (DRAGAN-NS and GAN-GP) only produced mode collapse on one dataset, see Figure~\ref{fig:mode_collapse_non_sat_gp}).
We also noticed that for certain learning rates, WGAN-GPs fail to learn the data distribution (Figure~\ref{fig:wgan_failure}). For the GAN-GP and DRAGAN-NS models, most hyperparameters produced samples of equal quality - the models are quite robust. We show samples from the GAN-GP, DRAGAN-NS and WGAN-GP models in Figures~\ref{fig:cifar_all_samples}, ~\ref{fig:celeba_all_samples} and ~\ref{fig:cmnist_all_samples}.

\subsection{Metrics}

We show that gradient penalties make non-saturating GANs more robust
to hyperparameter changes. For this, we report not only the best obtained
results, but rather a box plot of the obtained results showing the quartiles
obtained by each sweep, along with the top 10 best results explicitly shown in
the graph (note that for each model we tried 27 different hyperparameter settings, corresponding to 3 discriminator learning rates, 3 generator learning rates and 3 generator input sizes).
We report two Inception Score metrics for CIFAR-10, one using the
standard Inception network used when the metric was introduced \citep{improvedgan},
trained on the Imagenet dataset, as well as a VGG style network trained on CIFAR-10 (for details on the architecture, we refer the reader to \citet{alphagan}). We report the former to be compatible with existing literature, and the latter to obtain a more meaningful metric, since the network doing the evaluation was trained on the same dataset as the one we evaluate, hence the learned features will be more relevant for the task at hand. When reporting sample diversity, we subtract the average pairwise image similarity (as reported by MS-SSIM) computed as the mean of the similarity of every pair of images from 5 batches from the test set. Note that we can only apply this measure to CelebA, since for datasets such as CIFAR-10 different classes are represented by very different images, making this metric meaningless across class borders. Since our models are completely unsupervised, we do not compute the similarity across samples of the same class as in \citep{acgan}. The Inception Score and sample diversity metric results can be seen in Figure~\ref{fig:mssim_inception}. The results obtained using the Independent Wasserstein critic on all datasets can be found in Figure~\ref{fig:wcritics-all}.

\subsection{Key takeaways from real data experiments}

\begin{figure*}[!]
\begin{center}
\begin{subfigure}[Color MNIST]{
\includegraphics[width=0.3\textwidth]{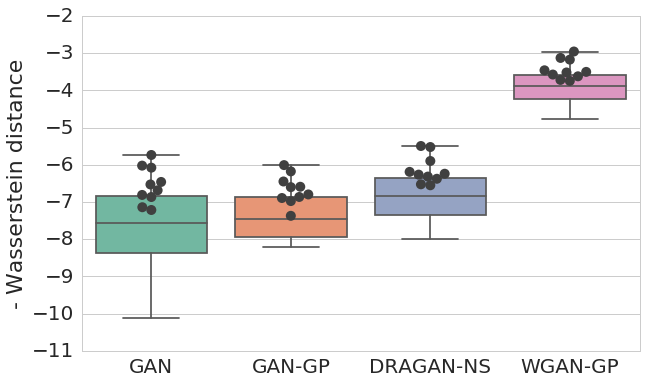}
\label{fig:wcritics-colormnist}
}\end{subfigure}
\begin{subfigure}[CelebA]{
\includegraphics[width=0.3\textwidth]{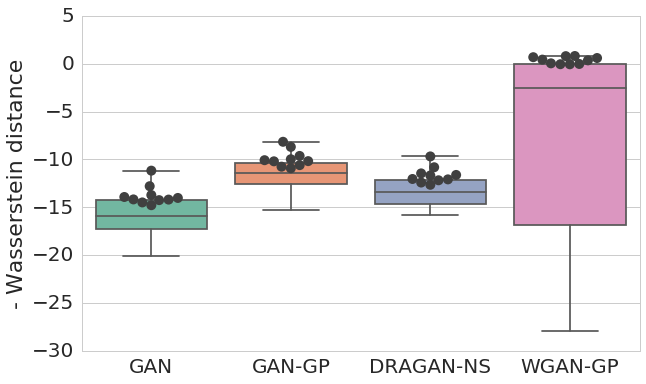}
\label{fig:wcritics-celeba}
}\end{subfigure}
\begin{subfigure}[CIFAR-10]{
\includegraphics[width=0.3\textwidth]{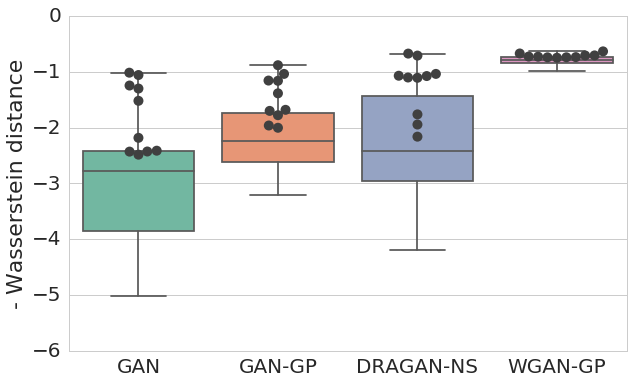}
\label{fig:wcritics-cifar}
}\end{subfigure}
\caption{{Negative Wasserstein distance estimated using an independent Wasserstein critic on the three datasets we evaluate on. The metric captures overfitting to the training data and low quality samples. Higher is better; the 10 black dots represent the results obtained with the 10 best hyperparameter settings.}
}
\label{fig:wcritics-all}
\end{center}
\vspace{-4mm}
\end{figure*}

\begin{figure*}[!]
\begin{center}
\begin{subfigure}[CelebA]{
\includegraphics[width=0.3\textwidth]{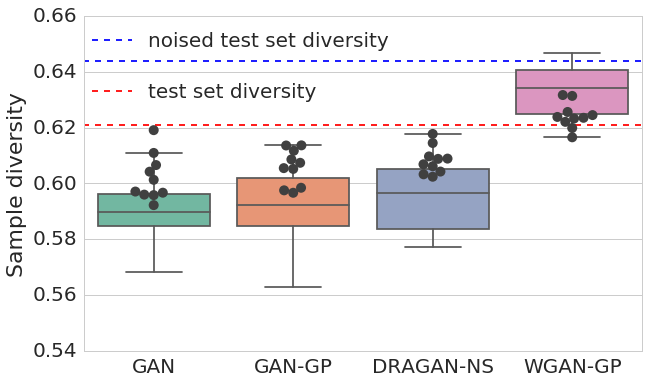}
\label{fig:mssim_inception:celeba}
}\end{subfigure}
\begin{subfigure}[Inception Score (ImageNet)]{
\includegraphics[width=0.3\textwidth]{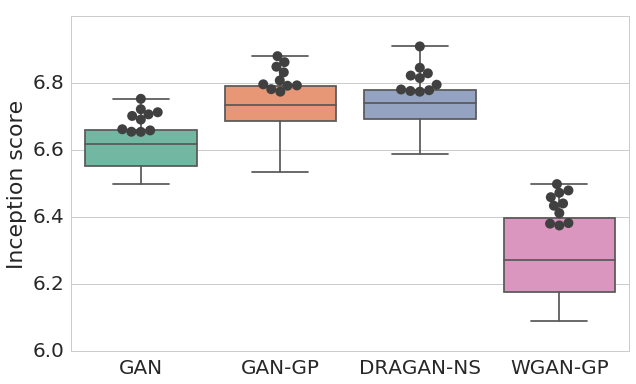}
\label{fig:mssim_inception:inceptionnet}
}\end{subfigure}
\begin{subfigure}[Inception Score (CIFAR)]{
\includegraphics[width=0.3\textwidth]{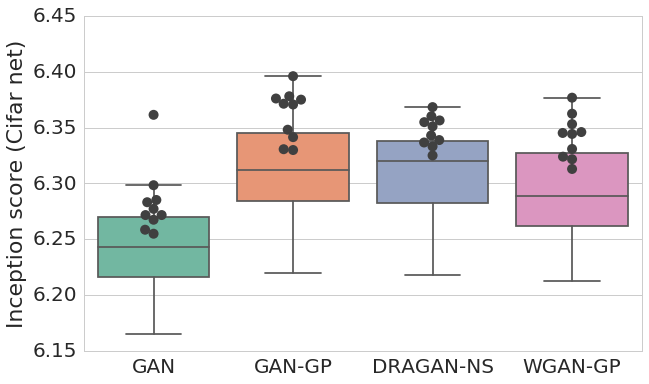}
\label{fig:mssim_inception:cifarnet}
}\end{subfigure}
\caption{
Left plot shows sample diversity results on CelebA. It is important to look at this measure relative to the measure on the test set: too much diversity can mean failure to capture the data distribution, too little is indicative of mode collapse. To illustrate this, we report the diversity obtained when adding normal noise with zero mean and 0.1 standard deviation to the test set: this results in more diversity than the original data. The black dots report the results closest to the reference values obtained on the test set by each model. Middle plot: Inception Score results on CIFAR-10. Right most plot shows Inception Score computed using a VGG style network trained on CIFAR-10.
As a reference benchmark, we also compute these scores using samples from test data split; diversity: 0.621, Inception Score: 11.25, Inception Score (VGG net trained on CIFAR-10): 9.18.
}
\label{fig:mssim_inception}
\end{center}
\vspace{-2mm}
\end{figure*}

When analyzing the results obtained by training non-saturating GANs
using gradient penalties (GAN-GP and DRAGAN-NS), we notice that:
\begin{itemize}
  \item Both gradient penalties help when training non-saturating GANs,
    by making the models more robust to hyperparameters.
  \item On CelebA, for various hyperparameter settings WGAN-GP fails to learn the data distribution and produces samples that do not look like faces (Figure~\ref{fig:wgan_failure}). This results in a higher sample diversity than the reference diversity obtained on the test set, as reported by our diversity metric - see Figure \ref{fig:mssim_inception:celeba} which compares sample diversity for the considered models across hyperparameters. The same figure shows that for most hyperparameter values, the WGAN-GP model produces higher diversity than the one obtained on the test set (indicating failure to capture the data distribution), while for most hyperparameters non-saturating GAN variants produce samples with lower diversity than that of the test set (indicating mode collapse). However, WGAN-GP is closer to the reference value for more hyperparameters, compared to the non-saturating GAN variants.
  \item Even if we are only interested in the best results (without looking across the hyperparameter sweep), we see that the gradient penalties tend to improve results for non-saturating GANs.
  \item The non-saturating GAN trained with gradient penalties
      produces better samples which give better Inception Scores, both when looking at the results obtained from the best set of hyperparameters and when looking at the entire sweep.
  \item While the non-saturating GAN variants are much faster to train than the WGAN-GP model (since we do only one discriminator update per generator update), they perform similarly to the WGAN-GP model. Thus, non-saturating GANs with penalties offer a better computation versus performance tradeoff.
  When we trained WGAN-GP models in which we update the discriminator only once per
  generator update, we noticed a decrease in sample quality for all datasets, reflected by our reported metrics, as seen in Figure~\ref{fig:wgan-1-update}.
  \item When looking at the independent Wasserstein critic results, we see that the WGAN-GP models perform best on Color MNIST and CIFAR-10. However, on CelebA the Independent Wasserstein Critic can distinguish between validation data examples and samples from the model (see Figure~\ref{fig:wcritics-celeba}). This is consistent with what we have seen by examining samples: the hyperparameters which result in samples of reduced quality are the same with a reduced negative Wasserstein distance.
  \item The sample diversity metric and the Independent Wasserstein critic detect  mode collapse. When DRAGAN-NS collapses for two hyperparameter settings, the negative Wasserstein distance reported by the critic for these jobs is low, showing that the critic captures the difference in distributions, and the sample diversity reported for those settings is greatly reduced (Figure~\ref{fig:celeba_worst_results}).
\end{itemize}

\section{Discussion}

We have shown that viewing the training dynamics of GANs through the lens of the underlying divergence at optimality can be misleading. On low-dimensional synthetic problems, we showed that non-saturating GANs are able to learn the true data distribution where Jensen-Shannon divergence minimization would fail. We also showed that gradient penalty regularizers help improve the training dynamics and robustness of non-saturating GANs. It is worth noting that one of the gradient penalty regularizers was originally proposed for Wasserstein GANs, motivated by properties of the Wasserstein distance;  evaluating non-saturating GANs with similar gradient penalty regularizers helps disentangle the improvements arising from optimizing a different divergence (or distance) and the improvements from better training dynamics.
\\ %

\noindent \textbf{Comparison between explored gradient penalties:}
As described in Section~\ref{sec:grad_penalties}, we have evaluated two gradient
penalties on non-saturating GANs. We now turn our attention to the distinction
between the two gradient penalties. We have already noted that for a few hyperparameter settings, DRAGAN-NS produced samples with mode collapse, while the GAN-GP model did not. By looking at the resulting metrics, we note that there is no clear winner between the two types of gradient penalties. To assess whether the two penalties have a different regularization effect, we also tried applying both (with a gradient penalty coefficient of 10 for both, or of 5 for both), but that did not result in better models. This could be because the two penalties have a very similar effect, or due to optimization considerations (they might conflict with each other).
 \\ %

\noindent \textbf{Other gradient penalties:}
Besides the gradient penalties explored in this work, several other regularizers have been proposed for stabilizing GAN training.
 \citet{fgan} proposed a gradient penalty aiming to smooth the discriminator of $f$-GANs (including the minimax GAN),  which we refer to as $f$-GAN-GP,
inspired by \citet{sonderby2016amortised} and \citet{arjovsky2017towards}. Their gradient penalty is different from the ones explored here; specifically, their gradient penalty is weighted by the square of the discriminator's probability of real for each data instance and the penalty is applied to data and samples (no noise is added).
In Fisher-GAN \citep{fishergan}, an equality constraint that is added on the
magnitude of the output of the discriminator on data as well as samples is
directly penalized, as opposed to the magnitude of the discriminator
gradients, as in WGAN-GP. Similar to WGAN-GP, the penalty was introduced in
the framework of integral probability metrics, but it can be directly applied
to other approaches to GAN training. Unlike WGAN-GP, Fisher GAN uses augmented Lagrangians to impose the equality
constraint, instead of a penalty method. To the best of
our knowledge, this has not been tried yet and we leave it for future work.
\\

The regularizers assessed in this work (the penalties proposed by DRAGAN and WGAN-GP), as well as others (such as $f$-GAN-GP and Fisher-GAN) are similar in spirit, but have been proposed from distinct theoretical considerations.
Future study of GAN regularizers will determine how these regularizers interact, and help us understand the mechanism by which they stabilize GAN training and motivate new approaches.

\subsection*{Acknowledgements} We thank Ivo Danihelka and Jascha Sohl-Dickstein for helpful feedback and discussions.

\clearpage
\newpage

\bibliography{resources}

\newpage
\appendix
\input{appendix.tex}

\end{document}

%% file: math_commands.tex
\def\eqref#1{equation~\ref{#1}}

\def\1{\bm{1}}

\def\vx{{\bm{x}}}

\def\vz{{\bm{z}}}
\newif\ifMIT
\MITfalse
\ifMIT

\def\gamma{{\boldmath{\vgamma}}}

\else

\def\vgamma{{\bm{\gamma}}}

\fi

\ifMIT

\else

\fi

\DeclareMathAlphabet{\mathsfit}{\encodingdefault}{\sfdefault}{m}{sl}
\SetMathAlphabet{\mathsfit}{bold}{\encodingdefault}{\sfdefault}{bx}{n}

\newcommand{\phatx}{p_{\hat{x}}}
\newcommand{\pnoise}{p_{\rm{noise}}}
\newcommand{\pdata}{p_{\rm{data}}}

\newcommand{\pmodel}{p_{\rm{model}}}

\newcommand{\E}{\mathbb{E}}

\newcommand{\R}{\mathbb{R}}

\ifMIT
\else

\fi

\DeclareMathOperator{\Tr}{Tr}

%% file: appendix.tex
\section{Results}
\subsection{Synthetic Experiments}

We present here more detailed results for our synthetic experiments.
\begin{figure}[ht]
\centering
\includegraphics[width=13cm]{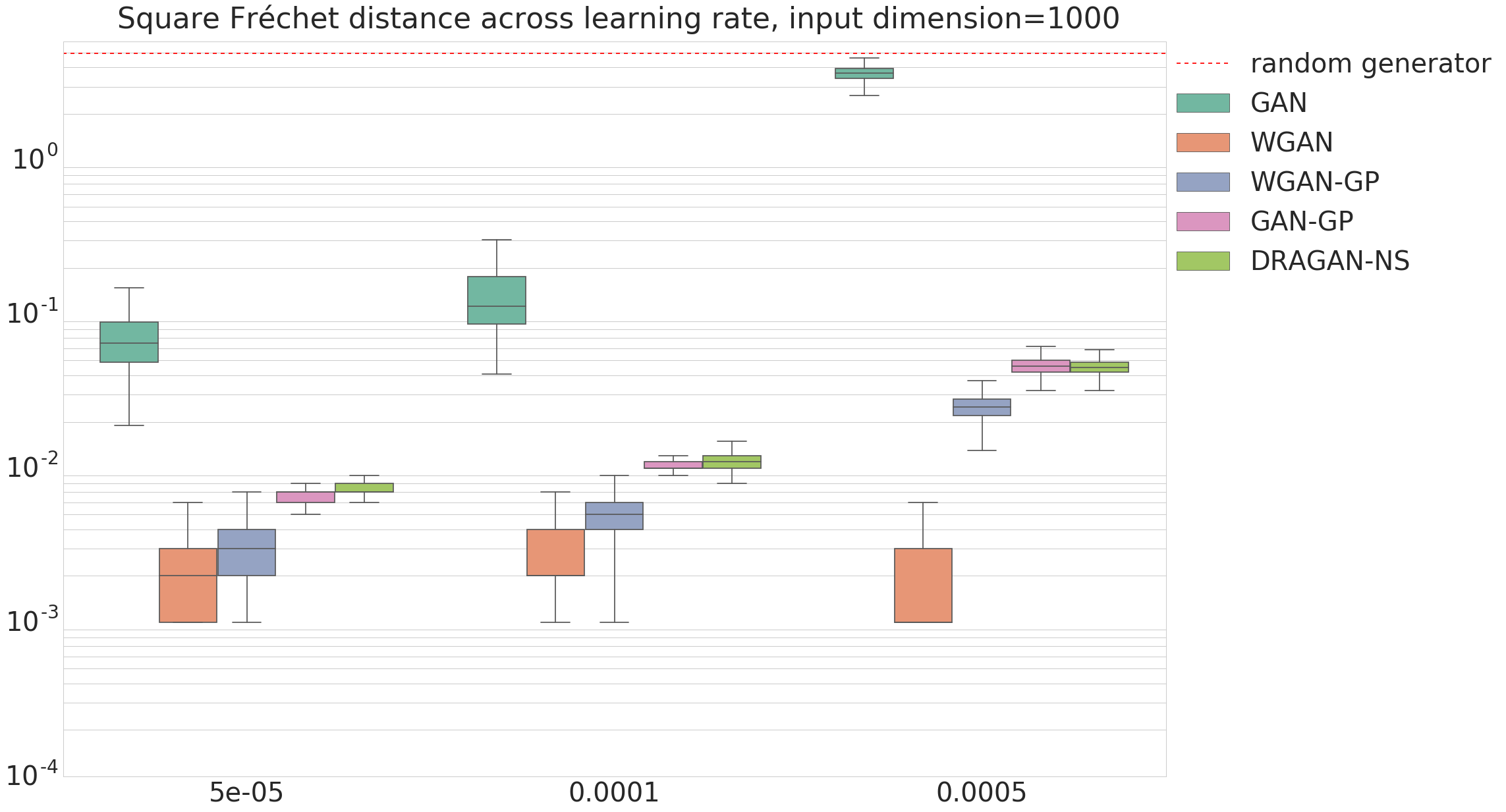} \\
\caption{\textbf{Synthetic Experiment 1.} The square {F}r\'{e}chet distance between the generated Gaussian parameters and true Gaussian parameters for different GAN variants, when varying the learning rate while keeping the input dimension fixed. Results averaged over 1000 runs. Lower values are better.}
\label{fig:hyper_exp1_lr}
\end{figure}

\begin{figure}[ht]
\centering
\includegraphics[width=13cm]{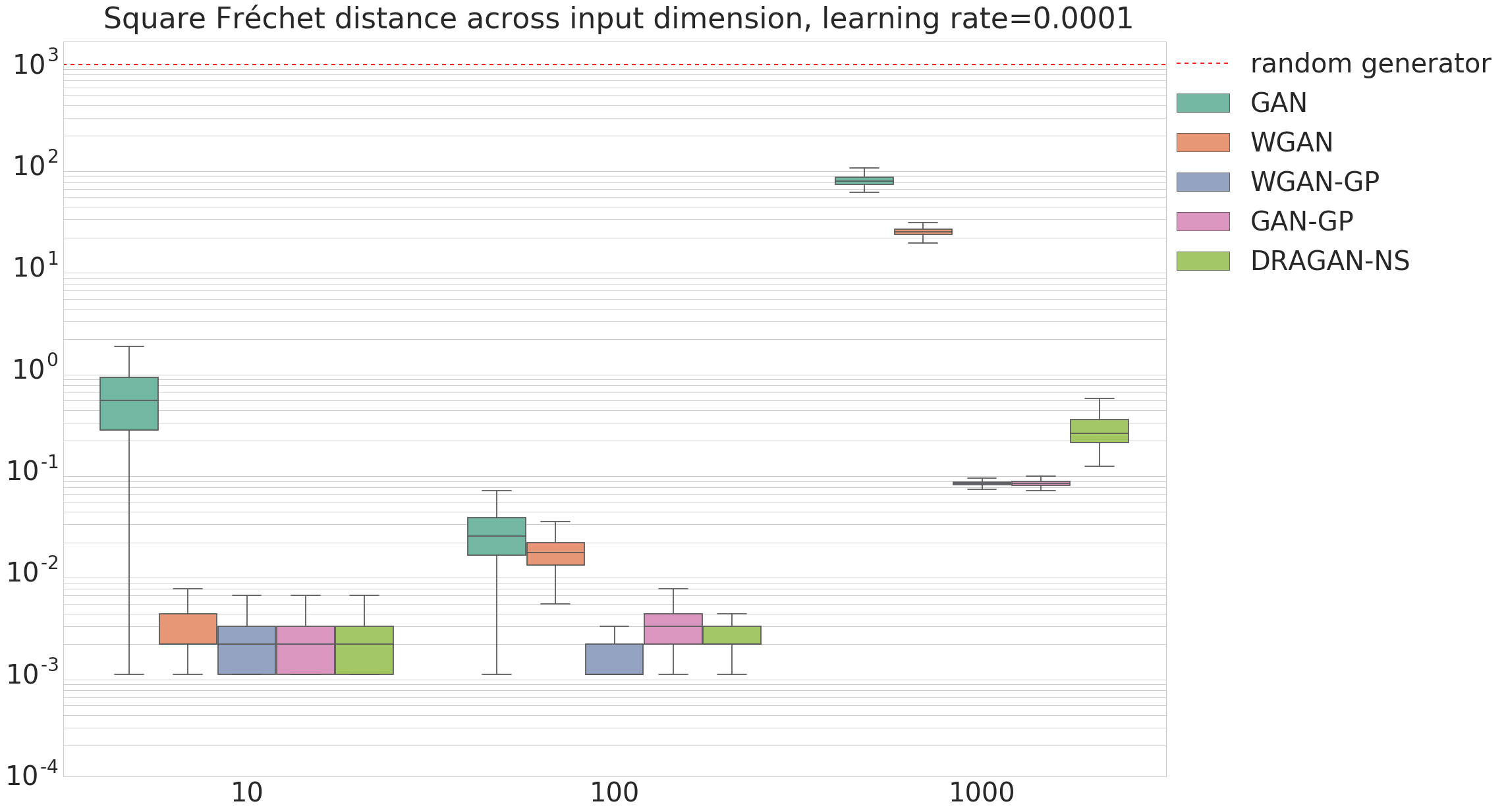} \\
\caption{\textbf{Synthetic Experiment 2.} The square {F}r\'{e}chet distance between the generated Gaussian parameters and true Gaussian parameters for different GAN variants, when varying the learning rate while keeping the input dimension fixed. Results averaged over 1000 runs. Lower values are better.}
\label{fig:hyper_exp2_input_dims}
\end{figure}

\subsection{The effect of the number of discriminator updates on GAN and GAN-GP}
\label{sec:synth_updates}

In this section we assess the affects of varying the discriminator update count per generator update. We notice that using 100 discriminator updates per generator update results in a bad distribution fit for the non saturating GAN. GAN-GP scales better with the number of discriminator updates but increasing the number of discriminator updates does not always result in a closer match to the true distribution for this model either.

\begin{figure}[ht]
\centering
\includegraphics[width=13cm]{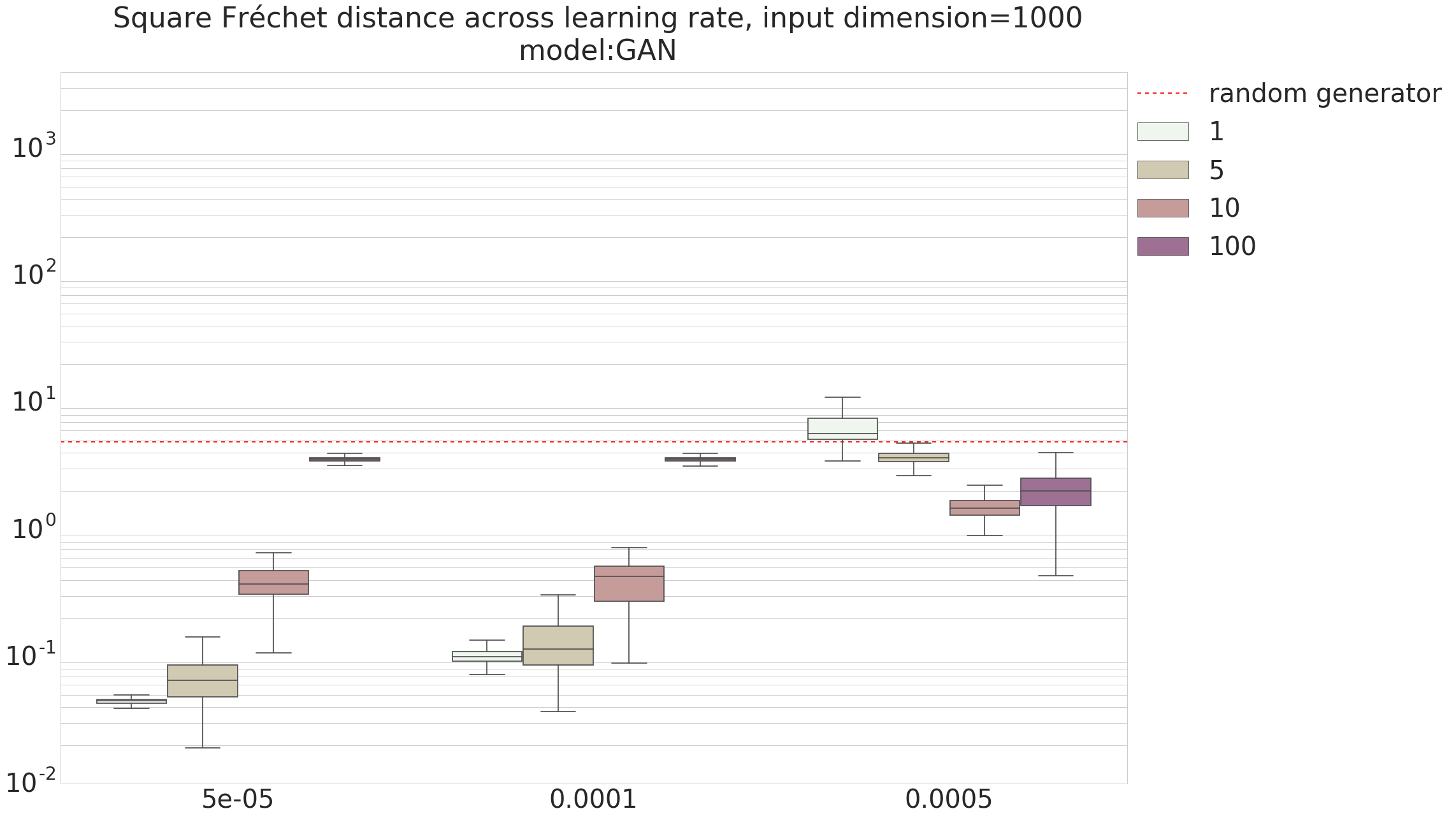} \\
\caption{\textbf{Synthetic Experiment 1.} The square {F}r\'{e}chet distance between the generated Gaussian parameters and true Gaussian parameters for different number of discriminator updates when training non saturating GANs, with varying the learning rates. Results averaged over 1000 runs. Lower values are better.}
\label{fig:hyper_exp1_lr_gan_update}
\end{figure}

\begin{figure}[ht]
\centering
\includegraphics[width=13cm]{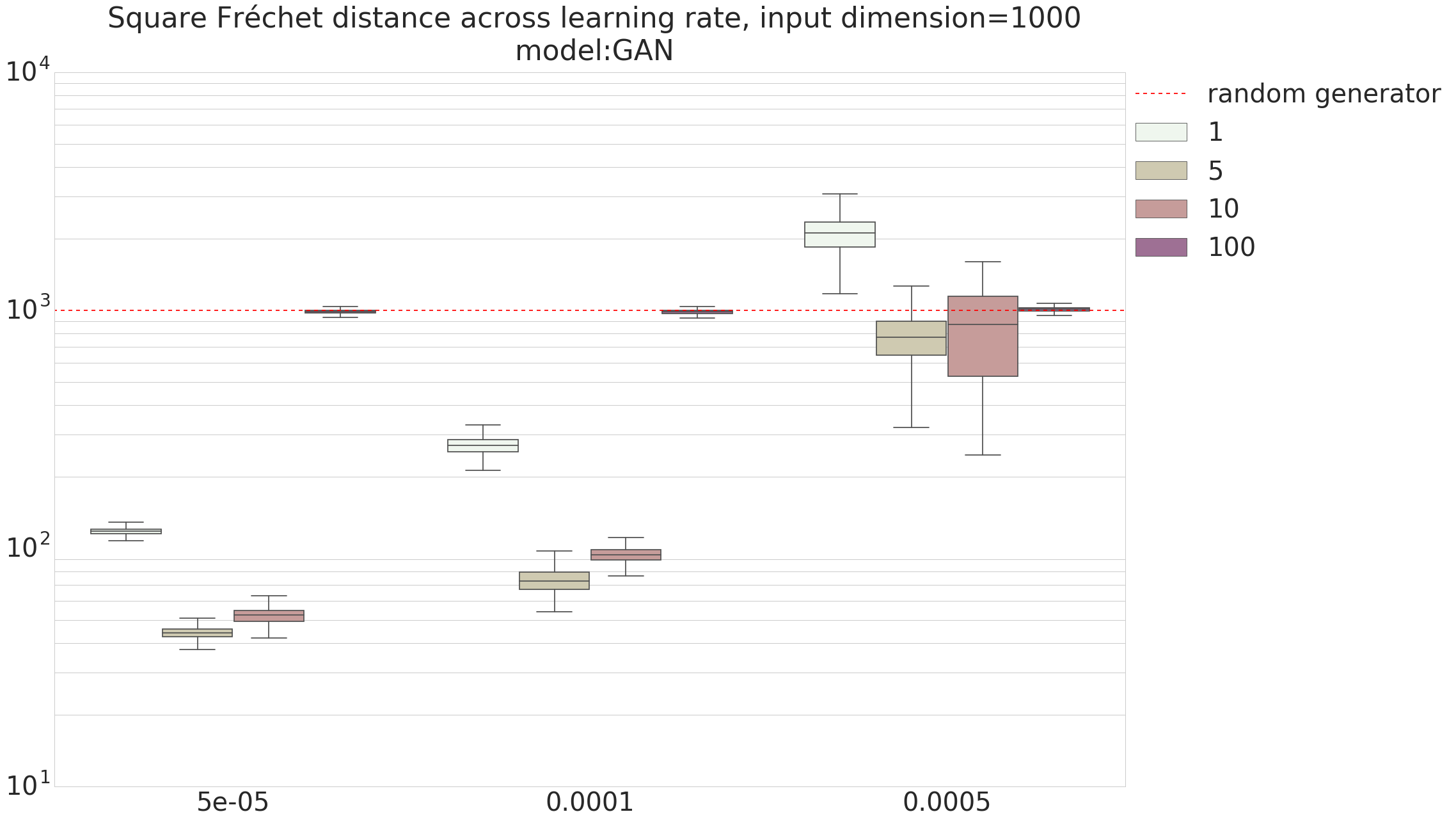} \\
\caption{\textbf{Synthetic Experiment 2.} The square {F}r\'{e}chet distance between the generated Gaussian parameters and true Gaussian parameters for different number of discriminator updates when training non saturating GANs, with varying the learning rates. Results averaged over 1000 runs. Lower values are better.}
\label{fig:hyper_exp2_lr_gan_update}
\end{figure}

\begin{figure}[ht]
\centering
\includegraphics[width=13cm]{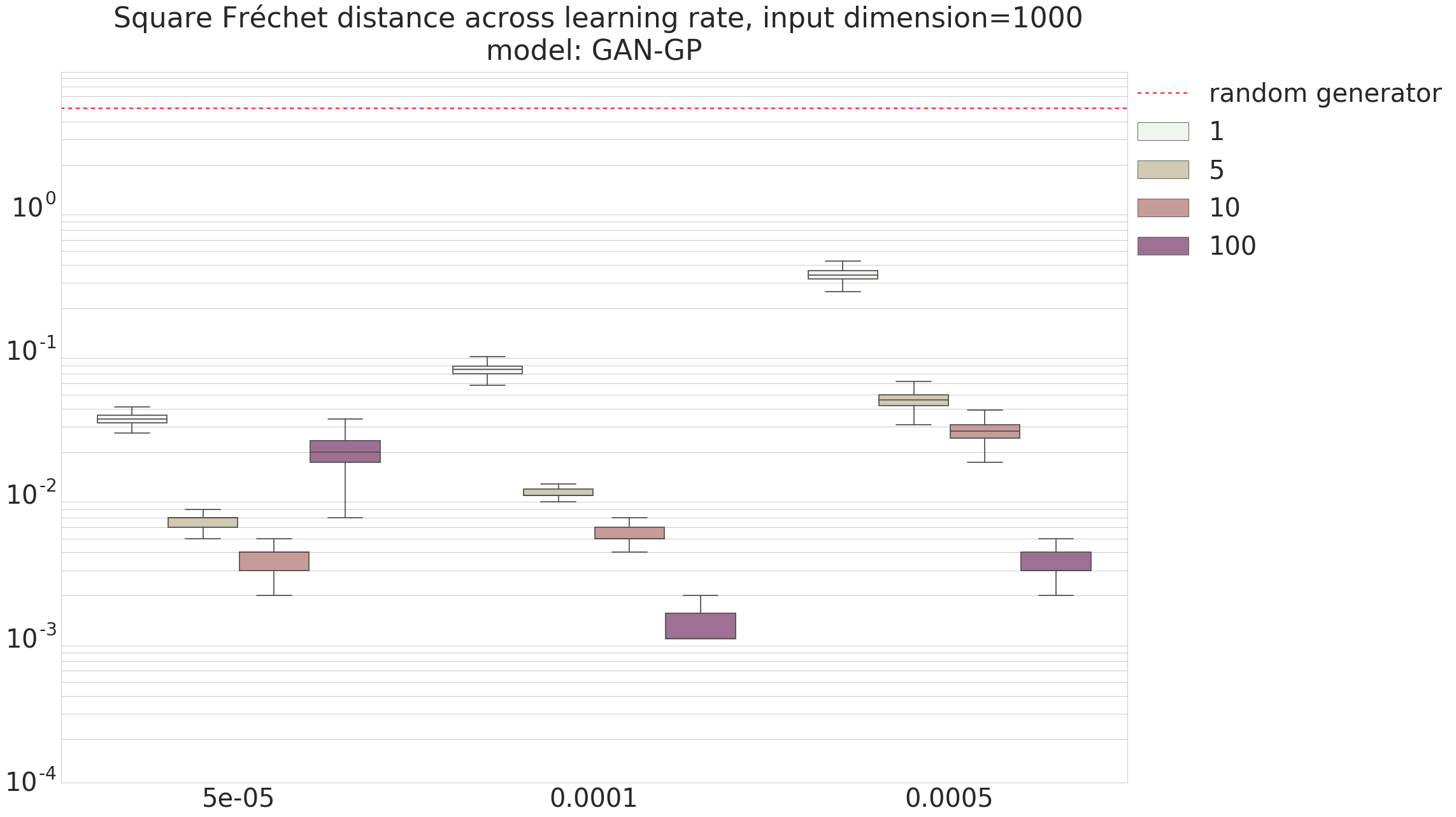} \\
\caption{\textbf{Synthetic Experiment 1.} The square {F}r\'{e}chet distance between the generated Gaussian parameters and true Gaussian parameters for different number of discriminator updates when training GAN-GP, with varying the learning rates. Results averaged over 1000 runs. Lower values are better.}
\label{fig:hyper_exp1_lr_gan_gp_update}
\end{figure}

\begin{figure}[ht]
\centering
\includegraphics[width=13cm]{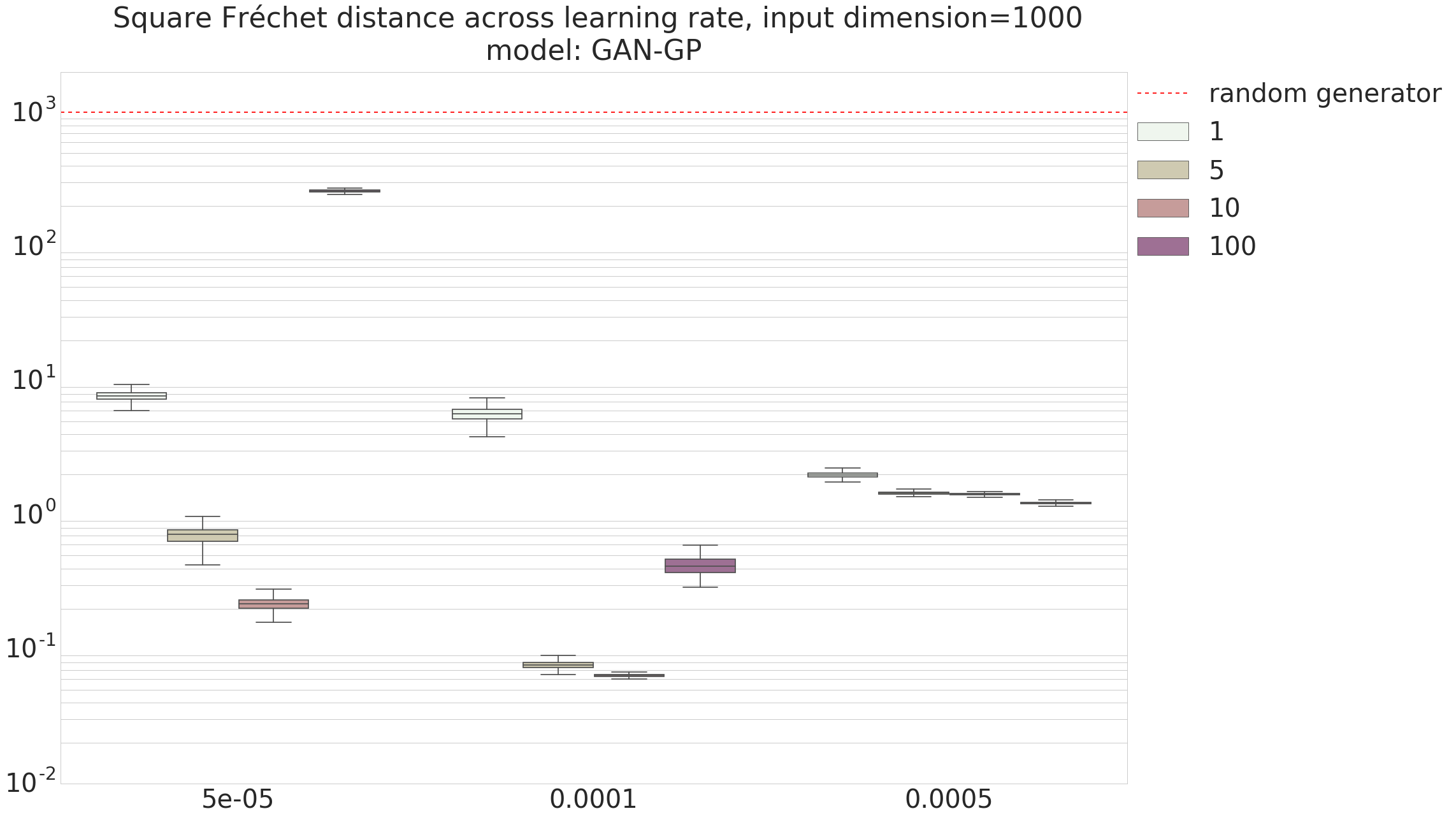} \\
\caption{\textbf{Synthetic Experiment 2.} The square {F}r\'{e}chet distance between the generated Gaussian parameters and true Gaussian parameters for different number of discriminator updates when training GAN-GP, with varying the learning rates. Results averaged over 1000 runs. Lower values are better.}
\label{fig:hyper_exp2_lr_gan_gp_update}
\end{figure}

\subsection{Real Data Experiments}
We present here generated samples and other evaluation metrics on real data.

\begin{figure*}[ht]
\begin{center}
\begin{subfigure}[Color MNIST]{
\includegraphics[width=0.3\textwidth]{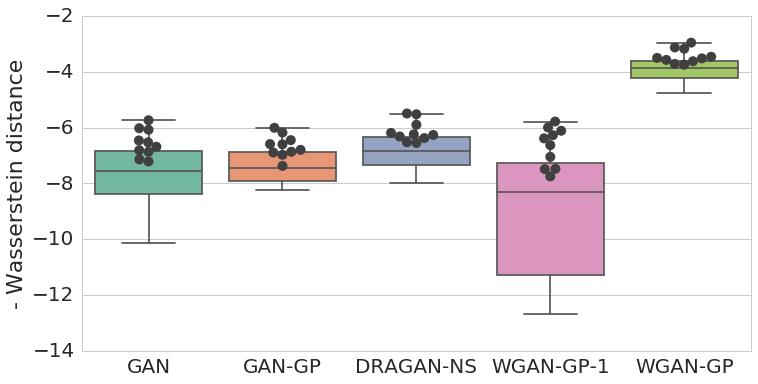}
\label{fig:wgan-1-update:wgan-mnist}
}\end{subfigure}
\begin{subfigure}[Inception Score (ImageNet)]{
\includegraphics[width=0.3\textwidth]{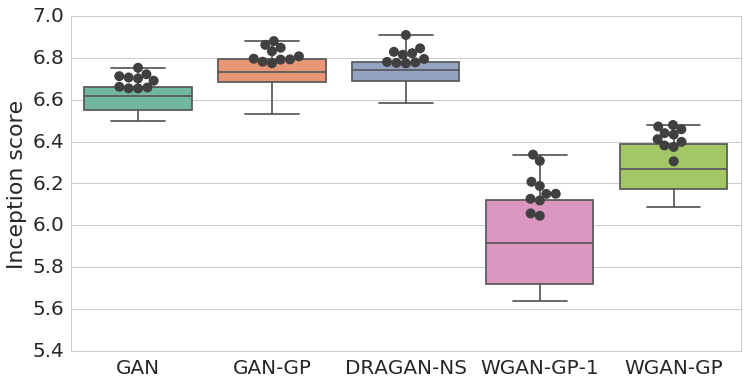}
\label{fig:wgan-1-update:inceptionnet}
}\end{subfigure}
\begin{subfigure}[Inception Score (CIFAR)]{
\includegraphics[width=0.3\textwidth]{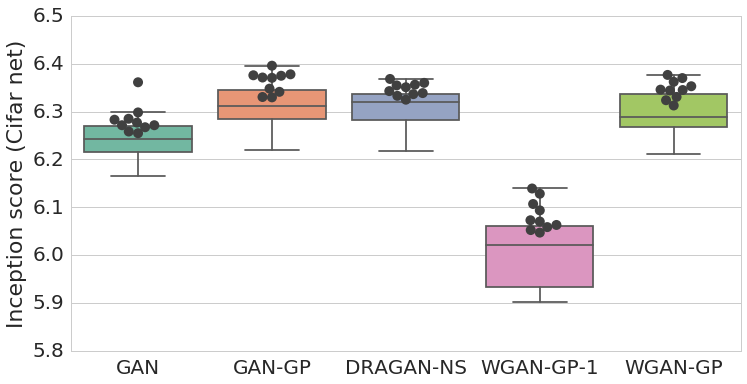}
\label{fig:wgan-1-update:cifarnet}
}\end{subfigure}
\caption{
Comparison across models when doing one update for the discriminator in Wasserstein GAN (WGAN-GP-1). The reduced performance in consistent with the observed decrease in sample quality when examining results. Inception Score results obtained on the test set: with Imagenet trained classifier: 11.25, With CIFAR-10 trained classifier: 9.18. Higher is better; the 10 black dots represent the results obtained with the 10 best hyperparameter settings.
}
\label{fig:wgan-1-update}
\end{center}
\vspace{-2mm}
\end{figure*}

\begin{figure*}[ht]
\begin{center}
\begin{subfigure}[CelebA- Sample diversity]{
\includegraphics[width=0.43\textwidth]{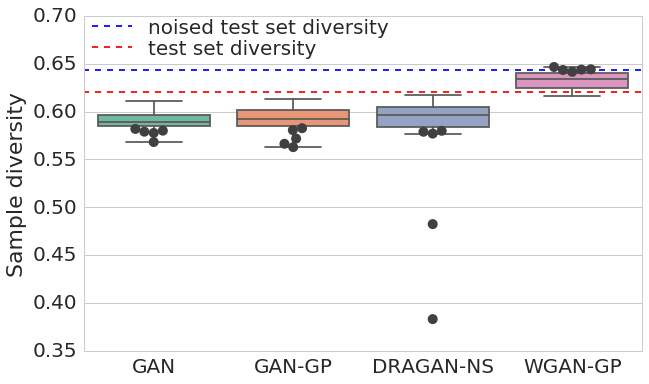}
\label{fig:worse_celeba_ms_ssim}
}\end{subfigure}
\begin{subfigure}[CelebA- Negative estimated Wasserstein distance]{
\includegraphics[width=0.43\textwidth]{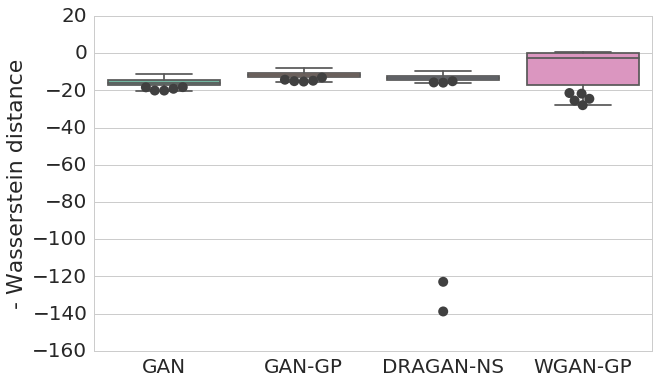}
\label{fig:worse_celeba_wgan}
}\end{subfigure}
\caption{
The metrics employed are able to capture mode collapse. Looking at the 5 worst values (the black dots) in a hyperparameter sweep according to sample diversity and negative Wasserstein distance as estimated by an Independent Wasserstein critic, we see that these metrics are able to capture the two examples of model collapse that we have seen when training DRGAN-NS on CelebA, as shown in Figure~\ref{fig:mode_collapse_non_sat_gp}. For sample diversity, the worst results are computed by the biggest absolute difference to the reference point (test set diversity), while for negative Wasserstein distance the worst results are computed by choosing the lowest value.
}
\label{fig:celeba_worst_results}
\end{center}
\vspace{-2mm}
\end{figure*}

\begin{figure*}[ht]
\begin{center}
\includegraphics[width=0.24\textwidth]{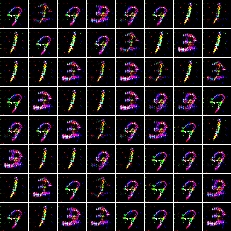} \addhspacesmall
\caption{Examples of mode collapse obtained for some hyperparameter settings with non-saturating GAN.}
\label{fig:mode_collapse_standard_gan}
\end{center}
\end{figure*}

\begin{figure*}[ht]
\begin{center}
\includegraphics[width=0.24\textwidth]{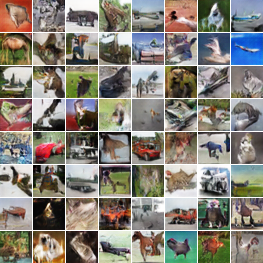} \addhspacesmall
\includegraphics[width=0.24\textwidth]{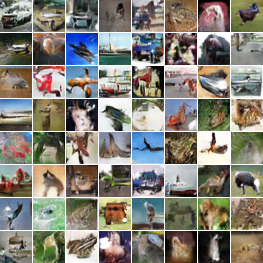} \addhspacesmall
\includegraphics[width=0.24\textwidth]{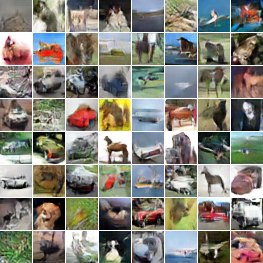} \addhspacesmall
\caption{CIFAR-10 samples obtained from the GAN-GP, DRAGAN-NS, and WGAN-GP models.}
\label{fig:cifar_all_samples}
\end{center}
\end{figure*}

\begin{figure*}[ht]
\begin{center}
\includegraphics[width=0.32\textwidth]{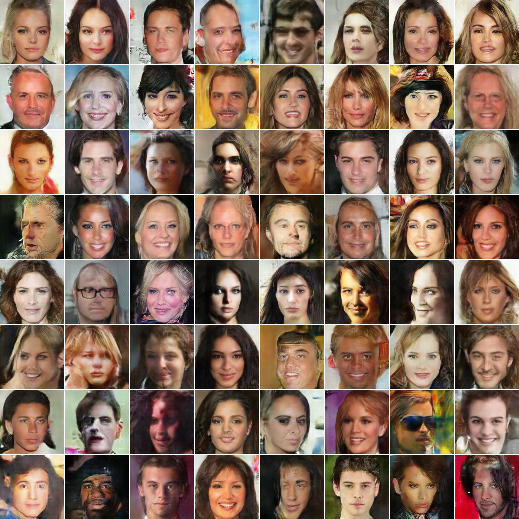} \addhspacesmall
\includegraphics[width=0.32\textwidth]{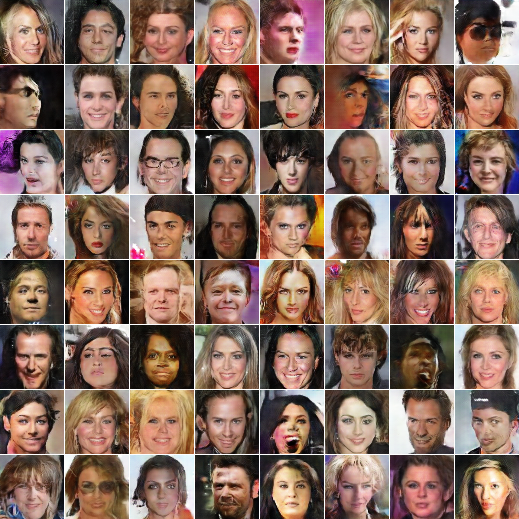} \addhspacesmall
\includegraphics[width=0.32\textwidth]{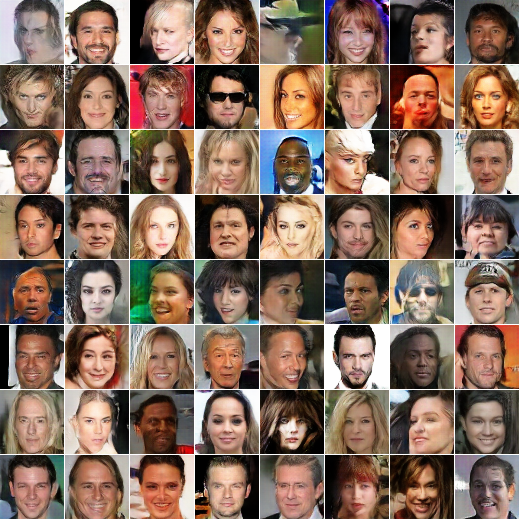} \addhspacesmall
\caption{CelebA samples obtained from the GAN-GP, DRAGAN-NS, and WGAN-GP models.}
\label{fig:celeba_all_samples}
\end{center}
\end{figure*}

\begin{figure*}[ht]
\begin{center}
\includegraphics[width=0.24\textwidth]{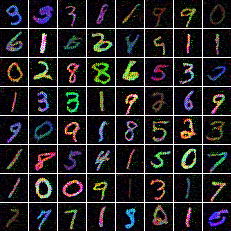} \addhspacesmall
\includegraphics[width=0.24\textwidth]{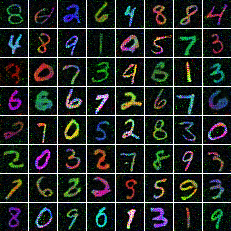} \addhspacesmall
\includegraphics[width=0.24\textwidth]{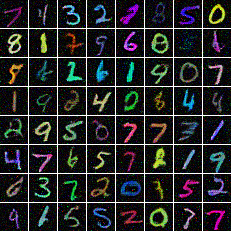} \addhspacesmall
\caption{CMNIST samples obtained from the GAN-GP, DRAGAN-NS, and WGAN-GP models.}
\label{fig:cmnist_all_samples}
\end{center}
\end{figure*}

\begin{figure*}[ht]
\begin{center}
\includegraphics[width=0.32\textwidth]{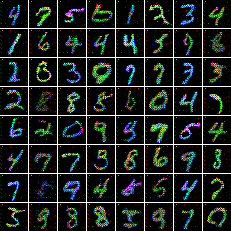} \addhspacesmall
\includegraphics[width=0.32\textwidth]{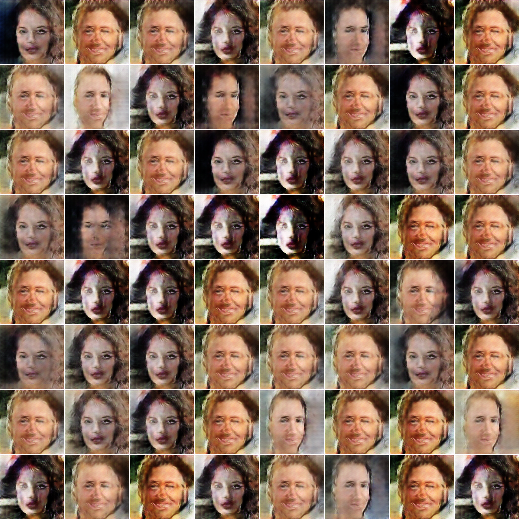} \addhspacesmall
\includegraphics[width=0.32\textwidth]{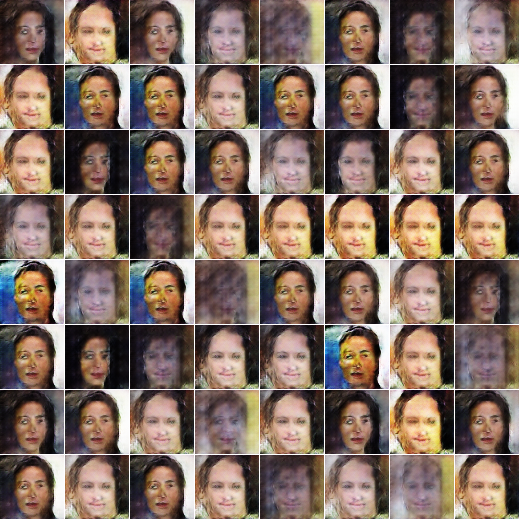} \addhspacesmall
\caption{Mode collapse when adding gradient penalties to non-saturating GANs. GAN-GP only had two instances of mode collapse, namely color mode collapse on Color-MNIST (left), while DRAGAN-NS only had two instances of mode collapse, which ocurred when trained on CelebA (right and middle).}
\label{fig:mode_collapse_non_sat_gp}
\end{center}
\end{figure*}

\begin{figure*}[ht]
\begin{center}
\includegraphics[width=0.32\textwidth]{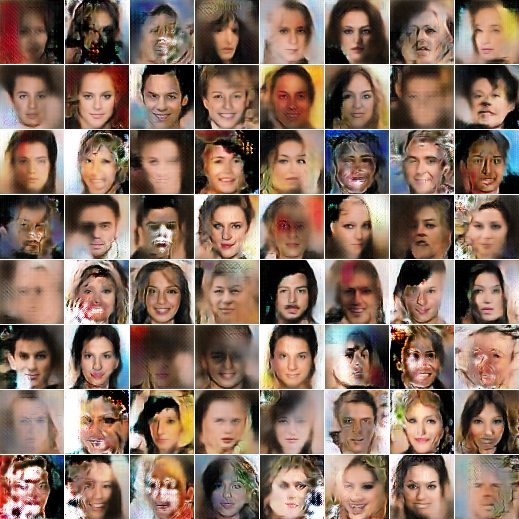} \addhspacesmall
\includegraphics[width=0.32\textwidth]{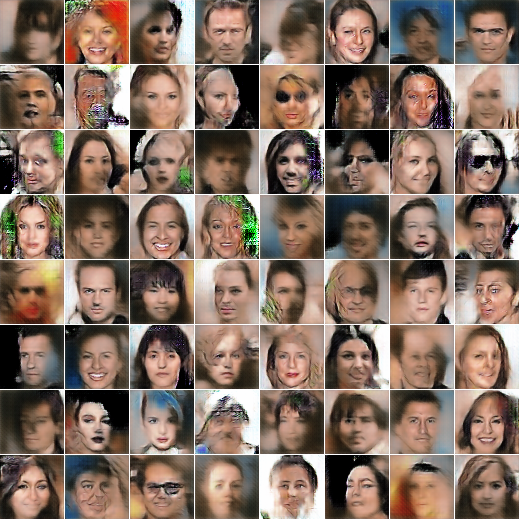} \addhspacesmall
\includegraphics[width=0.32\textwidth]{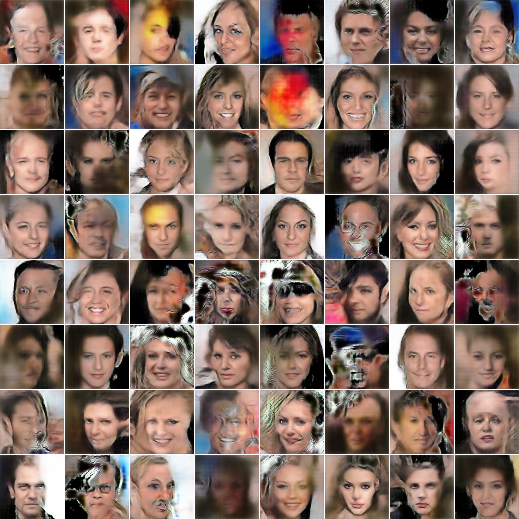} \addhspacesmall
\caption{Examples of failure to capture the data distribution with WGAN-GP. The model puts too much mass around the data distribution when trained on the CelebA dataset.}
\label{fig:wgan_failure}
\end{center}
\end{figure*}